\documentclass[10pt,twocolumn,letterpaper]{article}

\usepackage[pagenumbers]{cvpr} 

\usepackage{booktabs}
\usepackage{multirow}
\usepackage{graphicx}
\usepackage{array}

\definecolor{cvprblue}{rgb}{0.21,0.49,0.74}
\usepackage[pagebackref,breaklinks,colorlinks,allcolors=cvprblue]{hyperref}

\usepackage{xcolor}
\usepackage{xspace}
\newcommand{\name}{RoadSafe365\xspace}
\renewcommand\paragraph{\@startsection{paragraph}{4}{\z@}%
  {0.5ex \@plus 0.2ex \@minus 0.1ex}%
  {-0.5em}%
  {\normalfont\normalsize\bfseries}}
\makeatother

\def\paperID{-} 
\def\confName{CVPR}
\def\confYear{2026}

\usepackage{pifont}
\usepackage{tocloft}
\usepackage[toc,page,header]{appendix}
\usepackage{adjustbox}
\usepackage{minitoc}

\renewcommand \thepart{}
\renewcommand \partname{}

\title{Understanding Real-World Traffic Safety through RoadSafe365 Benchmark}

\author{\fontsize{11.5pt}{\baselineskip}\selectfont
    Xinyu Liu\textsuperscript{1}, Darryl Jacob\textsuperscript{1}, Yuxin Liu\textsuperscript{2}, Xinsong Du\textsuperscript{3}, Muchao Ye\textsuperscript{4}, Bolei Zhou\textsuperscript{2}, Pan He\textsuperscript{1\textdagger}\\[0.5mm]\fontsize{11pt}{\baselineskip}\selectfont
    \textsuperscript{1}Auburn University~~\textsuperscript{2}University of California, Los Angeles~~\textsuperscript{3}Harvard Medical School~~\textsuperscript{4}The University of Iowa~~\\
    \textsuperscript{\textdagger}Correspondence to: pan.he@auburn.edu
    \vspace{-1.25mm}
    }

\begin{document}

\maketitle
\doparttoc 
\faketableofcontents

\begin{abstract}
Although recent traffic benchmarks have advanced multimodal data analysis, they generally lack systematic evaluation aligned with official safety standards. To fill this gap, we introduce \name, a large-scale vision-language benchmark that supports fine-grained analysis of traffic safety from extensive and diverse real-world video data collections. Unlike prior works that focus primarily on coarse accident identification, \name~is independently curated and systematically organized using a hierarchical taxonomy that refines and extends foundational definitions of \textit{crash}, \textit{incident}, and \textit{violation} to bridge official traffic safety standards with data-driven traffic understanding systems. \name provides rich attribute annotations across diverse traffic event types, environmental contexts, and interaction scenarios, yielding 36,196 annotated clips from both dashcam and surveillance cameras. Each clip is paired with multiple-choice question–answer sets, comprising 864K candidate options, 8.4K unique answers, and 36K detailed scene descriptions that collectively designed for vision–language understanding and reasoning. We establish strong baselines and observe consistent gains when fine-tuning on \name. Cross-domain experiments on both real and synthetic datasets further validate its effectiveness. Designed for large-scale training and standardized evaluation, \name provides a comprehensive benchmark to advance reproducible research in real-world traffic safety analysis.
\end{abstract}    

\section{Introduction}
\label{sec:intro}

Threats to traffic safety, such as crashes, incidents, and violations, remain a pressing issue, resulting in substantial losses of lives, property, and societal welfare. 
In 2024 alone, motor vehicle crashes in the United States led to an estimated 39,345 deaths and nearly 2.2 million injuries~\cite{nhtsa2025early}, underscoring the urgent need for effective prevention to enhance public safety and traffic management~\cite{bao2021drive,fang2024abductive,moura2025nexar,you2020traffic}.
Although autonomous driving has advanced rapidly, models trained on routine driving datasets, despite their strong performance in general traffic understanding, struggle to recognize critical edge cases and rare failure scenarios such as traffic accidents and violations. A deeper understanding of traffic is crucial for anticipating risks and improving road safety. Encouragingly, recent advances in Multimodal Large Language Models (MLLMs), especially Vision-Language Models (VLMs), show strong ability to interpret complex scenes and generate detailed, context-aware explanations, outperforming conventional perception models that rely on black-box visual features~\cite{jain2024vcoder,shaw2025saferoute,keskar2025evaluating,wu2025trafficinternvl,sheng2025talk2traffic, fremont2019scenic} for the general-purpose traffic understanding.

\begin{figure*}[htb]
	\begin{minipage}[b]{1\textwidth}
		\centering		\includegraphics[width=\textwidth]{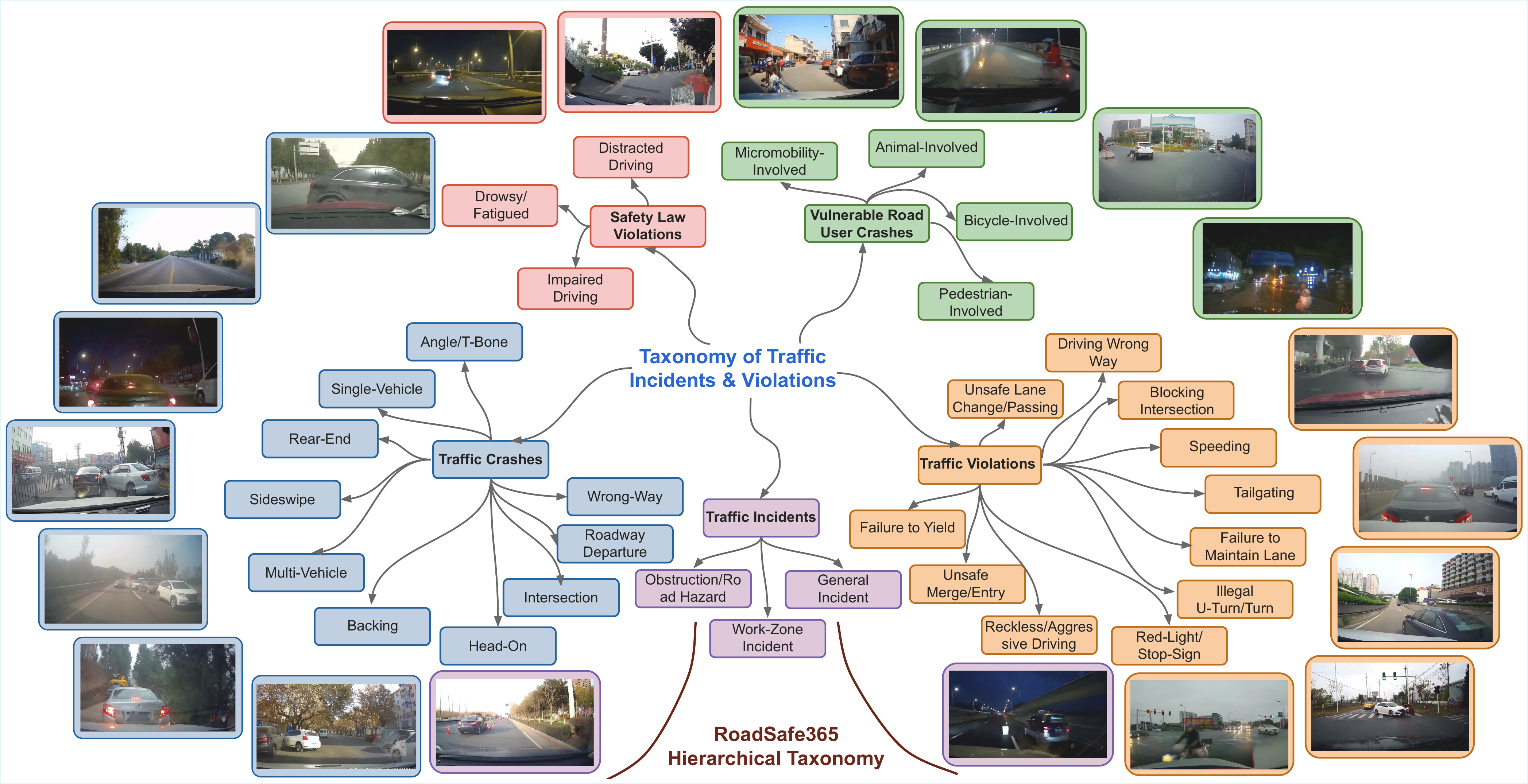}
	\end{minipage}
    \caption{Overview of the taxonomy for traffic safety understanding defined in the \name benchmark. 
    }
	\label{fig:taxonomy_tree} 
\end{figure*}

However, a major limitation remains: there is an absence of large-scale, high-quality datasets for training and evaluating vision–language models in traffic scenarios that require domain-specific, fine-grained safety understanding. Although several traffic accident datasets have been introduced to address this gap, they remain limited in scope: they often focus on specific aspects such as vulnerable road users (VRUs), perception, planning, or model robustness, but they lack systematically annotated datasets that support detailed accident causality reasoning and safety-critical evaluation of VLMs~\cite{kim2025vru,liu2025eventgpt,pan2024vlp,wang2025embodied,wang2025omnidrive,xie2025vlms}. More critically, existing datasets seldom organize traffic accidents and violations in a manner consistent with official safety standards, e.g., the definitions outlined in the Manual on Classification of
Motor Vehicle Traffic Crashes (ANSI D16) by the National Safety Council (NSC)~\cite{national1962manual}. Without such datasets, the gap between general-purpose and safety-critical tasks can lead to inaccurate or inconsistent responses, as VLMs, even advanced ones such as GPT-4o~\cite{hurst2024gpt}, Qwen2-VL~\cite{wang2024qwen2}, InternVL~\cite{chen2024internvl}, and Gemini-1.5-Pro~\cite{team2024gemini}, may misinterpret the visual content of traffic accidents. This is because accidents and violations stem from long-tailed, rare-event distributions that differ markedly from everyday traffic patterns. 

To address these gaps, we introduce \name, a large-scale vision–language benchmark for fine-grained traffic safety understanding built upon extensive real-world video data collected from public platforms across multiple countries and regions, forming a diverse and high-quality visual foundation. We propose a new hierarchical taxonomy (as shown in Fig.~\ref{fig:taxonomy_tree}) aligned with official safety standards and develop a rigorous annotation pipeline that produces comprehensive, human-verified annotations to support detailed reasoning tasks. This enables \name~to offer rich attribute annotations containing diverse event types, environmental conditions, and interaction scenarios, with 36,196 annotated clips captured from both dashcam and surveillance perspectives. Each clip is accompanied by multiple-choice Visual Question Answering (VQA) sets covering five safety-critical categories, comprising 864K candidate options, 8.4K unique answers, and 36K detailed scene descriptions, collectively enabling comprehensive vision–language understanding and reasoning. As a foundational benchmark for training and evaluating MLLMs, \name~aims to advance fine-grained accident understanding and generalizability in safety-critical scenarios. Our main contributions are summarized as the following:
\begin{itemize}
\item We introduce \name, a large-scale vision-language benchmark for traffic safety
understanding that includes 36,196 independently collected videos from both dashcam and surveillance perspectives.

\item We propose a hierarchical taxonomy that refines and extends foundational traffic safety concepts of \textit{crash}, \textit{incident}, and \textit{violation} to align official safety standards with data-driven fine-grained traffic understanding.  

\item Leveraging the new taxonomy in \name, we design a structured annotation pipeline that provides detailed QA attributes and narrative captions aligned with traffic safety standards. Comprehensive evaluations of state-of-the-art VLMs show that fine-tuning on \name~improves both VQA and captioning performance while generalizing to unseen accident datasets, including a public real-world benchmark and another new synthetic dataset \name-Synthetic. We aim to promote reproducible and systematic research in traffic safety.

\end{itemize}


\begin{figure*}[t]
	\begin{minipage}[b]{1\textwidth}
		\centering
		\includegraphics[width=0.96\textwidth]{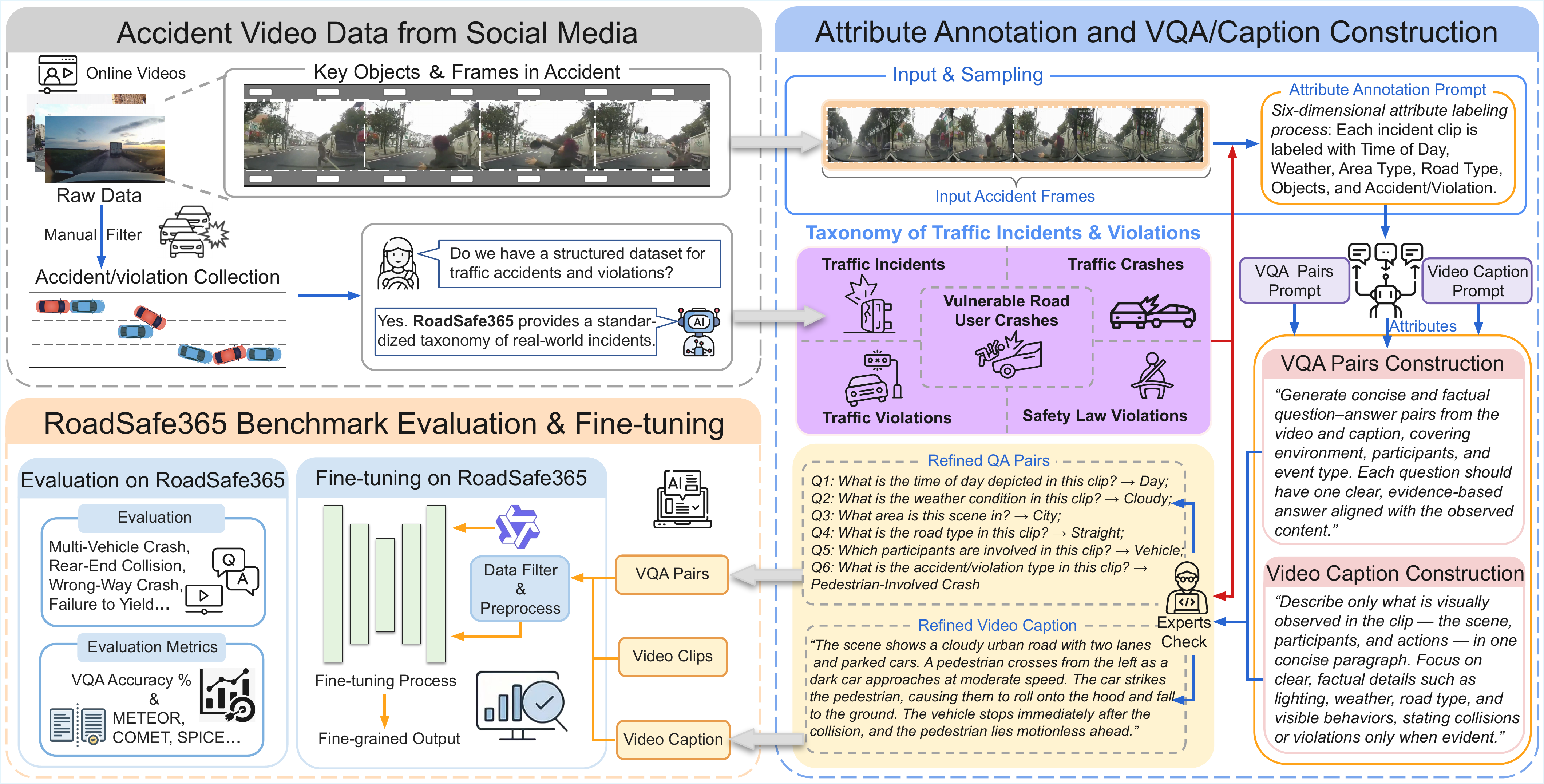}
	\end{minipage}
    \caption{Overview of \name: Collection, Annotation, Training, and Evaluation 
    }
	\label{fig:pipeline} 
\end{figure*}

\section{Related Work}
\label{Sec:Related Work}
\noindent \textbf{Accident and Driving Scene Understanding Datasets.} Recent works have introduced various datasets to support research on accident and driving understanding, 
which have incorporated language to support complex reasoning. 
We find that there is a lack of datasets in accident scene understanding. To illustrate,
MAPLM~\cite{cao2024maplm} provides a large-scale benchmark combining 2D images, 3D LiDAR, and HD map annotations focusing on map and scene-level reasoning, but ignores accident or violation analysis; MetaVQA~\cite{wang2025embodied} builds an embodied VQA benchmark for spatial and temporal reasoning in dynamic driving scenes, but it does not focus on detailed accident causality or standard safety-based categories; and NuPlanQA~\cite{park2025nuplanqa} explores multi-view and multimodal understanding in driving environments, but it mainly targets general perception instead of structured accident and violation reasoning.
Although datasets such as DADA-2000~\cite{fang2019dada}, MM-AU~\cite{bose2023mm}, and TAU-106K~\cite{zhou2025tau} offer real-world videos with causal annotations for accident scene understanding, they primarily aggregate publicly available videos with coarse labeling, limiting systematic analysis aligned with official safety standards. Also, datasets such as VRU-Accident~\cite{kim2025vru} and RoadSocial~\cite{parikh2025roadsocial} target specific scenarios, such as VRUs or road events from social media, and lack comprehensive coverage of diverse accidents. To overcome these limitations, we introduce \name, a large-scale dataset offering standardized and comprehensive annotations for diverse traffic accidents, enabling fine-grained safety analysis aligned with official traffic safety standards.



\noindent \textbf{Multimodal Large Language Models.} Traffic research, particularly in autonomous driving, has increasingly embraced MLLMs that integrate visual perception into LLMs, enabling textual reasoning over visual scenes. Example MLLMs applied to driving tasks include foundation models like InternVL~\cite{chen2024internvl}, Qwen-VL~\cite{wang2024qwen2}, and LLaVA~\cite{li2024llava}, and specialized models such as DriveVLM~\cite{sima2024drivelm} and LMDrive~\cite{shao2024lmdrive}. More recent frameworks include VLP~\cite{pan2024vlp}, a vision-language planning framework that uses language models to improve trajectory planning in autonomous vehicles, and DriveGPT4-V2~\cite{xu2025drivegpt4}, a  closed-loop driving pipeline integrating LLMs to enable language-guided decision-making and multi-stage reasoning.
Despite these advances, most works focus on normal driving scenes, with training data lacking diverse and realistic accident scenarios due to long-tailed, rare-event distributions—limiting their ability to reason about collisions, violations, and causal relationships.
To bridge this gap, we introduce our dataset to provide the foundation for developing more reliable and robust MLLMs in safety-critical environments.

\noindent \textbf{Reliable and Interpretable Accident Understanding.} Given the importance of reliability and interpretability for MLLMs in traffic understanding, researchers have developed specialized assessment frameworks to evaluate these aspects. To illustrate,
DriveBench~\cite{xie2025vlms} and AUTOTrust~\cite{xing2024autotrust} provide systematic frameworks to assess model trustworthiness and robustness under corruptions, measuring dimensions such as safety, uncertainty, and fairness; AutoDrive-QA~\cite{khalili2025autodrive} introduces a multiple-choice QA benchmark focused on driving scenarios, aiming to improve standardized evaluation of vision-language models; and DriveArena \cite{yang2025drivearena} proposes a closed-loop generative simulation platform that evaluates reasoning and intervention performance of autonomous driving models.
However, most existing studies emphasize benchmarking over large-scale, real-world data collection. While they provide valuable evaluation tools, they lack the detailed accident annotations essential for developing safety-critical models. Our \name fills this gap by serving as both a large-scale training source and a comprehensive benchmark, supporting the development of more reliable and interpretable models.



    \begin{figure*}[htb]
	\begin{minipage}[b]{1\textwidth}
		\centering
		\includegraphics[width=0.99\textwidth]{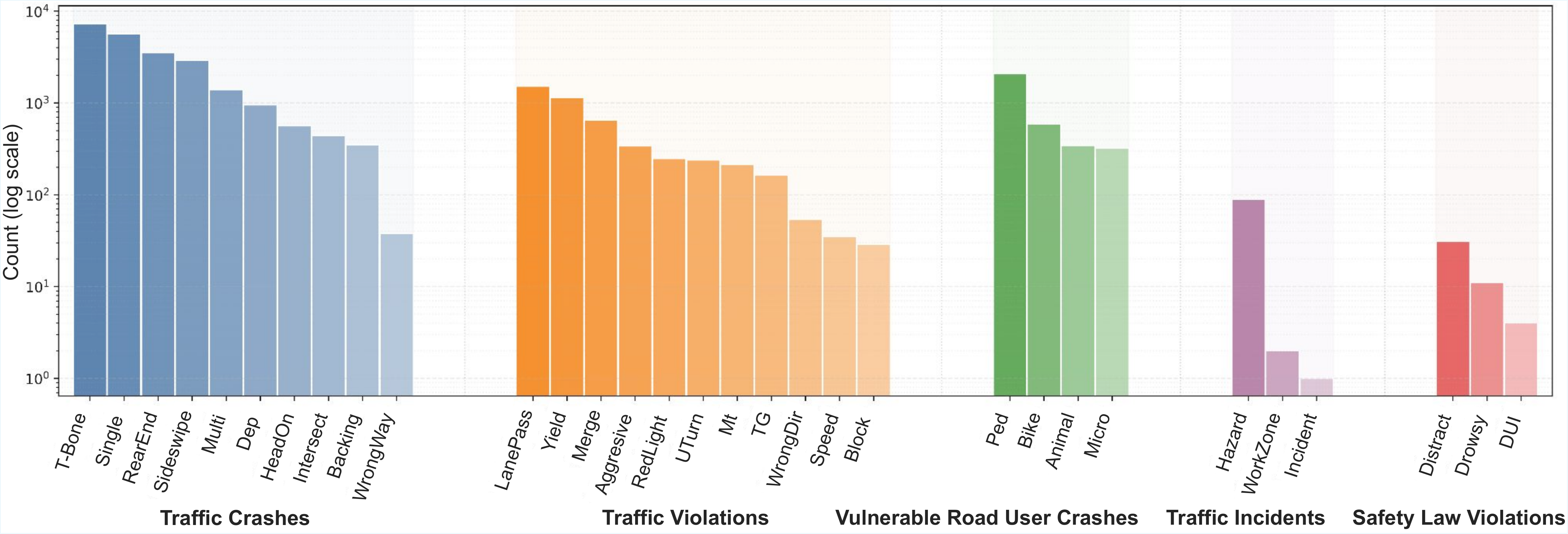}
	\end{minipage}
    \caption{\name Annotation Taxonomy and Data Distribution. 
    (a) shows the distribution (log scale) of the five Level-1 categories in our taxonomy. (b) illustrates the Level-2 subcategories nested within each Level-1 category. 
}
	\label{fig:taxonomy} 
\end{figure*}

\section{\name}

\subsection{Video Data Collection and Preprocessing}
Understanding traffic safety is vital for enhancing public safety and effective traffic management. However, existing public datasets are often limited in scale, diversity, and visual quality. As Fig.~\ref{fig:pipeline} shows, to build a more comprehensive and realistic collection, we curated accident videos from Bilibili and X (formerly Twitter), two popular social platforms that host extensive dashcam and surveillance footage. Our data collection strategy entails identifying blogger channels focused on traffic scene understanding—many of which host extensive crash and violation footage—and complementing this with keyword-based searches to retrieve additional accident-related videos. This process facilitates large-scale acquisition of real-world traffic footage covering a wide range of weather, lighting, and traffic conditions. For each raw footage, we further segmented it into multiple short clips using CapCut~\cite{CapCut2025}, an efficient video editing tool, followed by manual verification to ensure that each segment clearly depicts a traffic accident or other safety-critical event. Irrelevant, duplicate, and low-quality clips, e.g., those with severe occlusion or unstable viewpoints, were removed to maintain high data quality and consistency. In total, we collected $36{,}196$ accident clips, each lasting around $10$–$20$ seconds, spanning urban intersections, highways, residential areas, and rural roads. 

\subsection{Taxonomy for Traffic Safety Understanding}
As discussed, prior works primarily focus on coarse identification of traffic accidents or violations, lacking alignment with official safety standards established by traffic authorities. In this paper, as shown in Fig.~\ref{fig:taxonomy_tree}, \name~introduces a hierarchical taxonomy inspired by and adapted from widely recognized traffic incident standards, following definitions established by the NSC~\cite{national1962manual} to promote uniformity and comparability in motor vehicle traffic accident statistics. Specifically, we organize each clip by its event type following a two-level hierarchical structure: 

Firstly, as illustrated in Fig.~\ref{fig:taxonomy}(a), we categorize all traffic events into five primary Level-1 classes: traffic crashes, traffic violations, VRU crashes, traffic incidents, and safety law violations. The rationale behind the Level-1 category design is to assess each event’s impact based on person type, injury severity, and property damage. This structure enables detailed reasoning about questions such as whether a person was involved in a crash inside or outside a vehicle, the extent of property damage, the number of vehicles in transport involved, and the role of each person at the time of the crash, which provides more  insights beyond existing benchmarks.


Secondly, each Level-1 category is further divided into multiple fine-grained Level-2 subcategories, as visualized in Fig.~\ref{fig:taxonomy}(b). For instance, \textit{Traffic Crashes} include collision types such as rear-end, angle/T-bone (T-Bone), sideswipe, single-vehicle (Single), and intersection-related crashes. \textit{Traffic Violations} capture risky driving behaviors such as unsafe lane changes or passing (LanePass), failure to yield (Yield), red-light violations (RedLight), aggressive driving, and tailgating (TG). \textit{VRU Crashes} involve incidents with pedestrians (Ped), cyclists (Bike), animals, and micromobility (Micro) users, while \textit{Traffic Incidents} cover non-collision events such as roadway obstructions (Hazard) and work-zone disruptions. Lastly, \textit{Safety Law Violations} consider human-factor offenses such as distracted, drowsy, or impaired (DUI) driving.

Grounded in official safety standards, our hierarchical taxonomy enables structured accident analysis and causal reasoning, while existing benchmarks only use coarse, unsystematic event categories~\cite{zhou2025tau,kim2025vru}.

\subsection{Annotation Pipeline}
To enable fine-grained analysis, we followed the proposed hierarchical
taxonomy and developed an annotation pipeline that integrates structured attributes, question–answer pairs, and descriptive captions, as shown in Fig.~\ref{fig:pipeline}. Each video is first converted into a short montage of key frames and then annotated to ensure both structural consistency and rich contextual details.

 \noindent \textbf{Attribute Annotation.} We extract six core attributes using Gemini-2.5-Pro~\cite{comanici2025gemini} and manually verify each for accuracy and consistency. These attributes include time of day, weather, area type, road type, objects involved, and accident/violation category. The accident/violation label follows our two-level hierarchical structure consisting of a coarse class (e.g., collision, violation, or unknown) and a fine class aligned with standardized traffic categories (e.g., rear-end collision, unsafe lane change, failure to yield). When sufficient evidence is unavailable, annotators assign the label unknown rather than inferring intent, ensuring objectivity and reliability across all data sources.

 \noindent \textbf{Descriptive Captions.} Each clip is also accompanied by a detailed textual description generated by GPT-4o~\cite{hurst2024gpt}, focusing exclusively on visually observable elements such as the scene context, participants, their actions and interactions, and the resulting outcomes. All generated captions were subsequently reviewed and refined by human annotators to ensure accuracy and clarity.

 \noindent \textbf{VQA Construction.} To support question–answering tasks, the verified attributes are transformed into multiple-choice questions. Each question includes one correct answer and three distractors sampled from semantically similar categories to reduce trivial guessing. For multi-label questions (e.g., objects involved), distractors are generated by slightly modifying the correct set (e.g., drop-one or add-one). For accident or violation questions, we employ a curated combination bank to ensure that all answer choices are plausible and domain-relevant. Both automated validation and manual review are applied to maintain consistency, accuracy, and overall annotation quality.

All prompt templates and implementation details for attribute generation, captioning, and VQA construction are provided in the Appendix for interested readers.

\subsection{Summary of \name~Statistics}
The \name~dataset comprises $36{,}196$ real-world traffic accident video clips, each annotated with structured attributes, multiple-choice question–answer pairs, and detailed descriptive captions. In total, it includes over 200K attribute labels, 36K dense captions, and more than 200K VQA pairs. The annotations offer diverse accident scenarios with fine-grained labels and detailed event descriptions that capture both spatial and temporal context across a range of scenarios.

\section{Experiments}

In this section, we evaluate several state-of-the-art MLLMs on the proposed \name~benchmark across four core tasks: VQA, Dense Captioning, Fine-tuning Evaluation, and Cross-Domain Generalization, to analyze their performance and generalization capabilities.

\subsection{Experimental Setup}
We evaluate both open-source and closed-source MLLMs on the \name~benchmark. These models include Qwen2.5-VL-7B~\cite{bai2025qwen2}, InternVL3-8B~\cite{chen2024internvl}, LLaVA-NeXT-Video-7B~\cite{li2024llavanext}, Gemini 2.5 Pro~\cite{comanici2025gemini}, and GPT-4o~\cite{hurst2024gpt}, covering a diverse range of model sizes and architectures. For fine-tuning, we adopt VAU-R1~\cite{zhu2025vau} as the base model, leveraging its reinforcement learning–based optimization framework for video anomaly understanding.


\subsection{Tasks and Evaluation Protocol}

\begin{table*}[t]
\centering
\caption{Comparison on the VQA task of \name. Columns report accuracy (\%) for each attribute (Time, Weather, Area, Road Type, Object, Accident/Violation). Best and second-best are marked in {\color[HTML]{FE0000} \textbf{red}} and {\color[HTML]{3531FF} \textbf{blue}}.}
\tiny
\resizebox{\textwidth}{!}{%
\begin{tabular}{@{}l c | ccccccc@{}}
\toprule
\multirow{2}{*}{Models} & \multirow{2}{*}{Year} &
\multicolumn{6}{c}{Attribute Accuracy (\%)} & \multirow{2}{*}{\textbf{Avg.}} \\
\cmidrule(l){3-8}
& & Time & Weather & Area & Road & Object & Acc./Vio. & \\
\midrule
LLaVA-OneVision-7B~\cite{li2024llava}      & 2024 &  92.70 &  69.89 &  74.94 &  39.61 &  95.69 &  39.51 &  68.39 \\
InternVL3-2B~\cite{chen2024internvl}            & 2025 &  88.44 &  {\color[HTML]{FE0000} \textbf{75.63}} &  73.43 &  66.31 &  23.54 &  37.36 &  60.12 \\
InternVL3-8B~\cite{chen2024internvl}            & 2025 &  89.77 &  {\color[HTML]{3531FF} \textbf{75.31}} &  75.08 &  {\color[HTML]{3531FF} \textbf{74.07}} &  99.17 &  46.31 &  {\color[HTML]{FE0000} \textbf{76.62}} \\
InternVL3.5-4B~\cite{wang2025internvl3}          & 2025 &  88.16 &  72.19 &  65.44 &  68.15 &  98.16 &  {\color[HTML]{3531FF} \textbf{47.91}} &  73.34 \\
InternVL3.5-8B~\cite{wang2025internvl3}          & 2025 &  78.89 &  72.51 &  69.99 &  72.79 &  96.24 &  35.61 &  70.01 \\
Qwen2.5-VL-3B~\cite{bai2025qwen2}           & 2025 &  87.47 &  65.63 &  68.15 &  51.68 &  0.50 &  37.40 &  51.14 \\
Qwen2.5-VL-7B~\cite{bai2025qwen2}           & 2025 &  {\color[HTML]{FE0000} \textbf{94.77}} &  66.73 &  68.98 &  65.17 &  35.70 &  38.04 &  61.57 \\
Qwen3-VL-4B~\cite{bai2025qwen2}             & 2025 &  {\color[HTML]{3531FF} \textbf{94.26}} &  72.97 &   77.56 &  {\color[HTML]{FE0000} \textbf{74.30}} &  54.43 &  47.59 &  70.52 \\
Qwen3-VL-8B~\cite{bai2025qwen2}             & 2025 &  83.85 &  74.62 &  76.09 &  73.47 &  69.11 &  39.88 &  69.50 \\
LLaVA-NeXT-Video-7B~\cite{li2024llavanext}     & 2024 &  63.42 &  53.60 &  70.90 &  44.75 &  {\color[HTML]{FE0000} \textbf{99.86}} &  25.01 &  59.25 \\
MiniCPM-V 4.5~\cite{yao2024minicpm}             & 2024 &  72.74 &  44.84 &  {\color[HTML]{FE0000} \textbf{79.76}} &  30.38 &  {\color[HTML]{3531FF} \textbf{99.72}} &  32.63 &  60.01 \\
Gemini 2.5-flash~\cite{comanici2025gemini}          & 2025 &  83.43 &  75.26 &  {\color[HTML]{3531FF}\textbf{78.89}} &  70.54 &  88.53 &  {\color[HTML]{FE0000}\textbf{51.45}} &  {\color[HTML]{3531FF}\textbf{74.68}} \\
\bottomrule
\end{tabular}%
}
\label{tab:vqa_attribute_results}
\end{table*}

\begin{table*}[ht!]
\centering
\caption{Comparison on the dense caption task of \name. The best and second performance are marked in {\color[HTML]{FE0000} \textbf{red}} and {\color[HTML]{3531FF} \textbf{blue}}.}
\resizebox{\textwidth}{!}{%
\begin{tabular}{@{}l|c|c|c|ccc|ccc|ccc@{}}
\toprule
\multirow{2}{*}{Models} & \multirow{2}{*}{COMET↑} & \multirow{2}{*}{METEOR↑} & \multirow{2}{*}{SPICE↑} & \multicolumn{3}{c|}{ROUGE-1↑} & \multicolumn{3}{c|}{ROUGE-2↑} & \multicolumn{3}{c}{ROUGE-L↑} \\ \cmidrule(l){5-13} 
                        &                           &                           &                           & P & R & F & P & R & F & P & R & F \\ \midrule
LLaVA-OneVision-7B~\cite{li2024llava}   & 0.641 & 0.236 & 0.167 & 0.386 & 0.403 & 0.388 & 0.103 & 0.108 & 0.104 & 0.228 & 0.238 & 0.229 \\
InternVL3-2B~\cite{chen2024internvl}          & 0.707 & 0.303 & 0.220 & 0.450 & 0.504 & 0.472 & 0.145 & 0.163 & 0.152 & 0.256 & 0.287 & 0.269 \\
InternVL3-8B~\cite{chen2024internvl}          & {\color[HTML]{3531FF} \textbf{0.737}} & {\color[HTML]{3531FF} \textbf{0.317}} & {\color[HTML]{3531FF} \textbf{0.260}} & {\color[HTML]{3531FF} \textbf{0.510}} & {\color[HTML]{3531FF} \textbf{0.509}} & {\color[HTML]{FE0000} \textbf{0.506}} & {\color[HTML]{3531FF} \textbf{0.186}} & 0.184 & {\color[HTML]{3531FF} \textbf{0.184}} & {\color[HTML]{3531FF} \textbf{0.299}} & {\color[HTML]{3531FF} \textbf{0.298}} & {\color[HTML]{3531FF} \textbf{0.297}} \\
InternVL3.5-4B~\cite{wang2025internvl3}        & 0.736 & {\color[HTML]{FE0000} \textbf{0.320}} & 0.248 & 0.491 & {\color[HTML]{FE0000} \textbf{0.520}} & {\color[HTML]{3531FF} \textbf{0.503}} & 0.176 & {\color[HTML]{FE0000} \textbf{0.186}} & 0.180 & 0.289 & {\color[HTML]{FE0000} \textbf{0.306}} & 0.295 \\
InternVL3.5-8B~\cite{wang2025internvl3}        & {\color[HTML]{FE0000} \textbf{0.741}} & {\color[HTML]{FE0000} \textbf{0.320}} & {\color[HTML]{FE0000} \textbf{0.261}} & {\color[HTML]{FE0000} \textbf{0.546}} & 0.503 & 0.521 & {\color[HTML]{FE0000} \textbf{0.201}} & {\color[HTML]{3531FF} \textbf{0.185}} & {\color[HTML]{FE0000} \textbf{0.192}} & {\color[HTML]{FE0000} \textbf{0.318}} & 0.293 & {\color[HTML]{FE0000} \textbf{0.304}} \\
Qwen2.5-VL-3B~\cite{bai2025qwen2}         & 0.717 & 0.266 & 0.218 & 0.465 & 0.436 & 0.445 & 0.132 & 0.123 & 0.126 & 0.261 & 0.244 & 0.250 \\
Qwen2.5-VL-7B~\cite{bai2025qwen2}         & 0.719 & 0.266 & 0.221 & 0.479 & 0.424 & 0.447 & 0.149 & 0.131 & 0.139 & 0.280 & 0.247 & 0.261 \\
Qwen3-VL-4B~\cite{bai2025qwen2}           & 0.728 & 0.287 & 0.225 & 0.453 & 0.473 & 0.461 & 0.136 & 0.142 & 0.138 & 0.248 & 0.259 & 0.252 \\
Qwen3-VL-8B~\cite{bai2025qwen2}           & 0.734 & 0.306 & 0.222 & 0.455 & 0.505 & 0.476 & 0.138 & 0.152 & 0.144 & 0.256 & 0.283 & 0.267 \\
LLaVA-NeXT-Video-7B~\cite{li2024llavanext}   & 0.667 & 0.240 & 0.176 & 0.408 & 0.408 & 0.404 & 0.122 & 0.121 & 0.120 & 0.255 & 0.257 & 0.253 \\
MiniCPM-V 4.5~\cite{yao2024minicpm}          & 0.655 & 0.274 & 0.205 & 0.411 & 0.443 & 0.425 & 0.115 & 0.123 & 0.118 & 0.244 & 0.263 & 0.252 \\ \bottomrule
\end{tabular}
}
\label{tab:caption_evaluation}
\end{table*}

\textbf{VQA Task.} Each video is accompanied by multiple QA items spanning six categories: Time, Weather, Area, Road Type, Object, and Accident. Each question includes four candidate answers, one correct and three distractors. Models are prompted to select the correct option, and accuracy is computed for each category and averaged across all categories. This task emphasizes scene understanding and categorical reasoning, assessing how well models interpret multi-attribute traffic contexts and safety-critical events.

\noindent \textbf{Dense Captioning Task.} 
For this task, models generate detailed natural-language descriptions of traffic accident videos, capturing key entities, motion, and interactions. The generated captions are compared with ground truth annotations using SPICE~\cite{anderson2016spice}, METEOR~\cite{banerjee2004meteor}, COMET~\cite{rei2020comet}, and ROUGE~\cite{lin2004rouge} metrics to measure event coverage and linguistic quality. The classic BLEU~\cite{papineni2002bleu} metric is excluded due to its insensitivity to semantic and contextual variations. 

 \noindent \textbf{Fine-tuning Task.} To assess domain adaptability, we fine-tune Qwen2.5-VL-7B using the VAU-R1 training and evaluation pipeline. The model is reinforcement fine-tuned on the \name~QA subset with supervised question–answer pairs, without any additional instruction tuning or multi-stage optimization. We then evaluate the fine-tuned model on the held-out RoadSafe365 test set, enabling a direct comparison between pre-trained and fine-tuned models.
\noindent \textbf{Training and Evaluation.} 
We divide the RoadSafe365 dataset into distinct training and test sets with no overlap, allocating two-thirds for training and one-third for testing. 
From the training pool, $6{,}000$ accident clips are sampled for fine-tuning, each paired with six attribute-based VQA questions, resulting in $36{,}000$ QA pairs with fixed answer options for consistency. 
The Qwen2.5-VL-7B model is fine-tuned using the GRPO-based VAU-R1 framework for 3 epochs with a global batch size of $32$ (batch size of $1$ per GPU with gradient accumulation of 8 across 4 GPUs) and a learning rate of \(2 \times 10^{-6}\). 
Gradient checkpointing and FlashAttention-2 are enabled to optimize memory efficiency. Due to computational constraints, we only conduct a pilot study using only a subset of the training data to validate the effectiveness of fine-tuning on \name. Future work is highly encouraged to extend it by utilizing the full training split of \name~for large-scale training.

During inference, each video is represented by eight key frames, and all models are evaluated on a held-out test set of $700$  clips. 
We intentionally select a relatively small test set to balance inference time and evaluation quality, making the setup accessible to researchers with limited  resources.


\begin{figure*}[t]
	\begin{minipage}[b]{1\textwidth}
		\centering
		\includegraphics[width=0.95\textwidth]{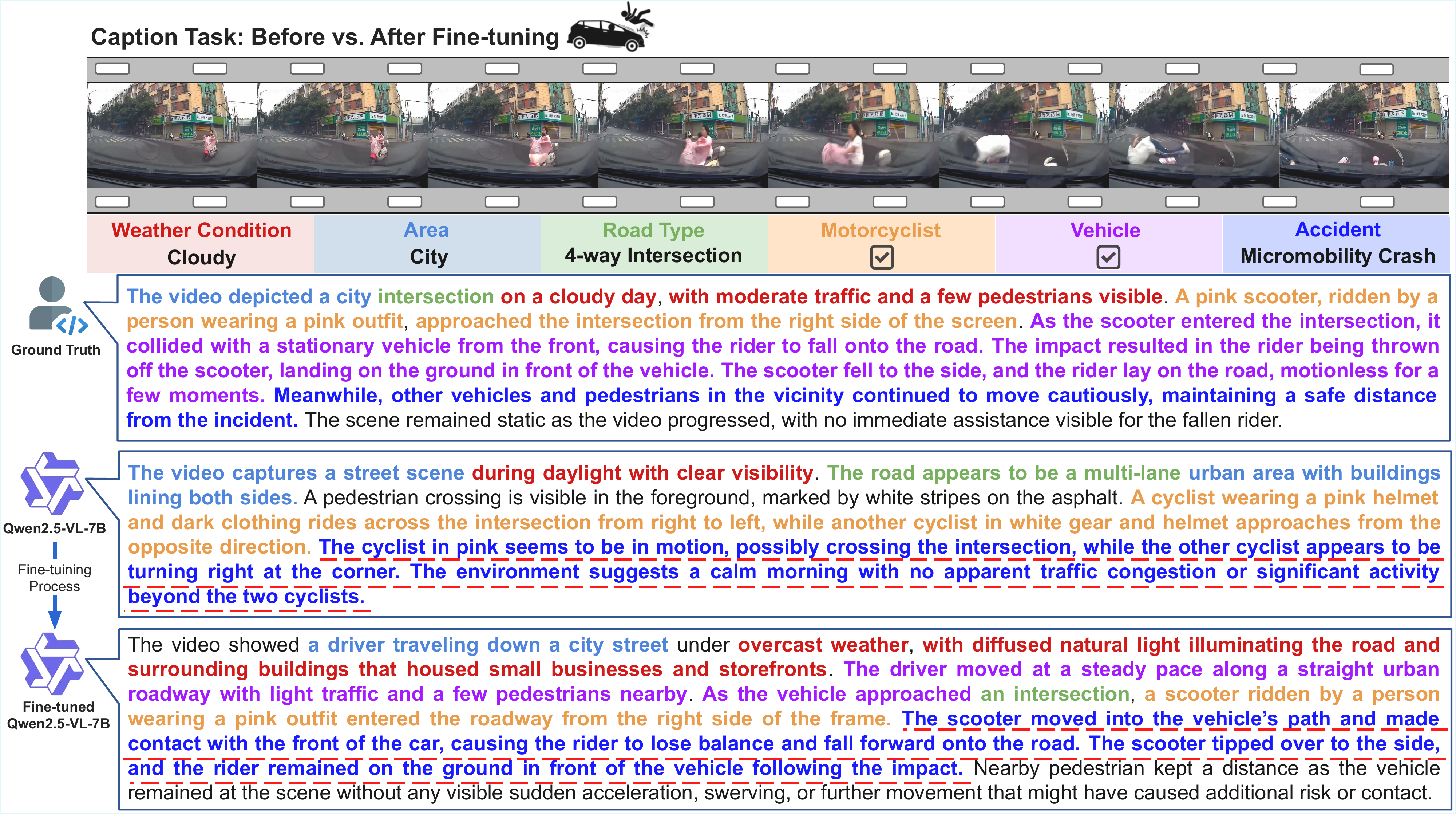}
	\end{minipage}
    \caption{Comparison of accident captions generated by Qwen2.5-VL-7B before and after fine-tuning on RoadSafe365.}
	\label{fig:finetune} 
\end{figure*}
\subsection{Main Results on Tasks}
\textbf{VQA Task.}  Table~\ref{tab:vqa_attribute_results} clearly demonstrates that \textit{InternVL3-8B} achieves the highest overall performance ($76.62\%$), followed by \textit{Gemini 2.5-flash} ($74.68\%$) and \textit{InternVL3.5-4B} ($73.34\%$). 
These models demonstrate strong performance on perception-focused attributes, particularly \textit{Time}, \textit{Weather}, and \textit{Area}, indicating robust visual grounding capabilities. In contrast, most models achieve considerably lower scores on \textit{Accident} and \textit{Violation}, which require causal and safety-related reasoning. 
Smaller open-source models, such as \textit{Qwen2.5-VL-3B} and \textit{InternVL3-2B}, struggle with these abstract reasoning categories, highlighting their limitations in causal understanding. 
Overall, perception-oriented attributes (Time, Weather, Area, Road, and Object) typically reach $70–95\%$ accuracy, whereas reasoning-oriented attributes (Acc./Vio.) often fall below $50\%$. 
This disparity suggests that while current VLMs can effectively recognize scene conditions, they still face challenges in inferring how and why incidents occur. Additional detailed evaluations of traffic events within the two-level hierarchical structure are provided in the appendix.

\noindent \textbf{Dense Captioning Task.} Table~\ref{tab:caption_evaluation} presents the dense captioning results on \name. 
\textit{InternVL3.5-8B} achieves the best overall performance across COMET, METEOR, SPICE, and ROUGE, demonstrating strong semantic alignment and well-structured descriptions. 
\textit{InternVL3-8B} ranks second on most metrics, indicating that larger model scales further enhance caption quality. 
Within the Qwen family, \textit{Qwen3-VL-8B} performs the best, generating fluent and detailed captions with the highest COMET and METEOR scores. 
In contrast, smaller open-source models exhibit clear limitations in capturing accident details and complex interactions. 
Overall, most models effectively describe scene context and visible actions but struggle with accident dynamics, causal reasoning, and multi-agent interactions. 



\begin{figure}[t]
    \centering
    \includegraphics[width=.4\textwidth]{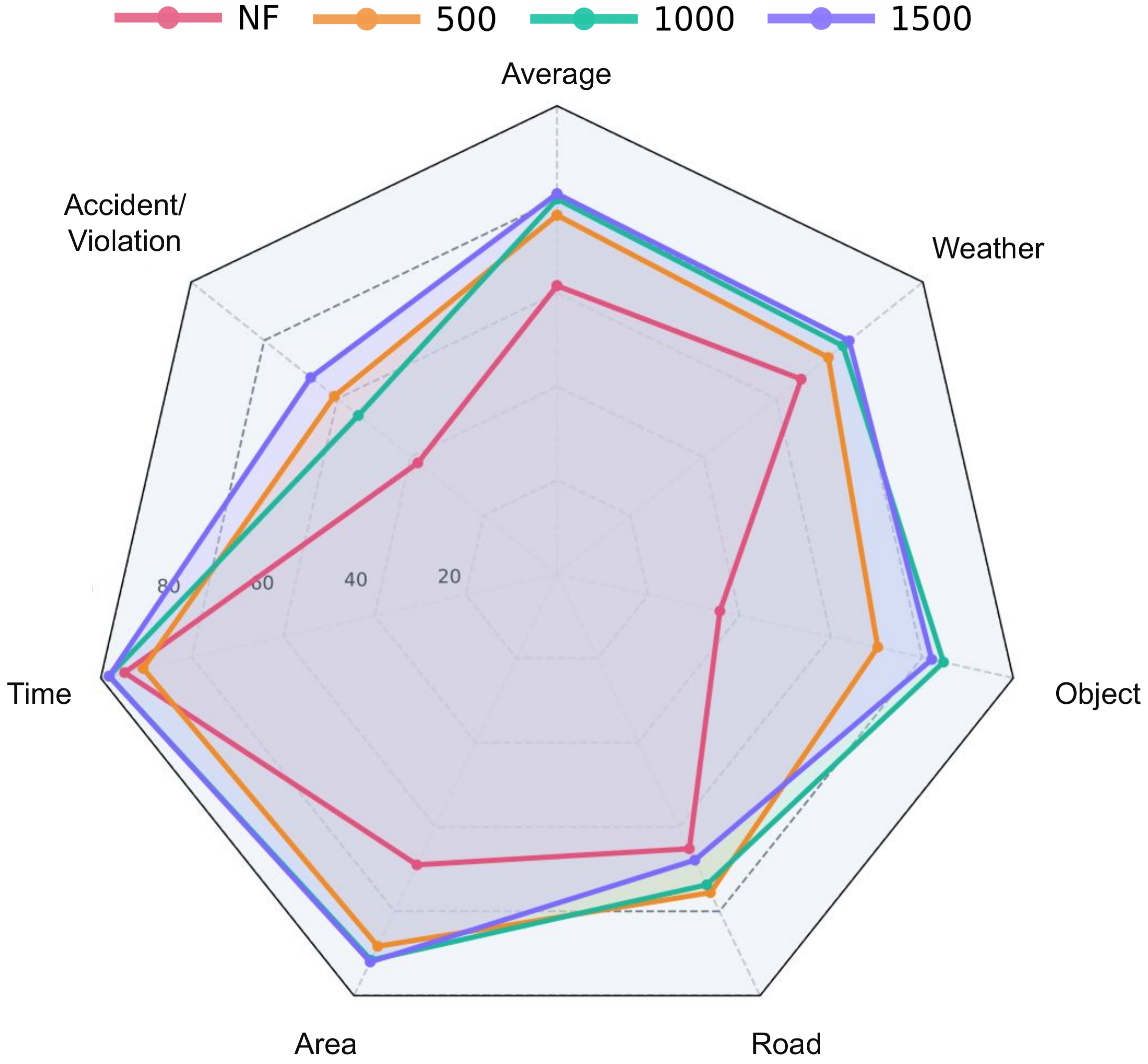}
    \caption{VQA performance of Qwen2.5-VL-7B fine-tuned on RoadSafe365 across different training iterations. NF denotes the original model without reinforcement fine-tuning.}
    \label{fig:lidar_chart}
\end{figure}

\noindent \textbf{Fine-Tuning Analysis.} We perform reinforcement fine-tuning on Qwen2.5-VL-7B using the VAU-R1 framework with VQA training samples from \name, and evaluate the resulting model on \name. As shown in Fig.~\ref{fig:lidar_chart}, More training iterations consistently improve the VQA performance of Qwen2.5-VL-7B across all categories. Notably, the Accident/Violation subcategory achieves a substantial gain—from $38.04\%$ to $67.29\%$---an absolute improvement of $29.25\%$, demonstrating a significant reduction in the gap for safety-critical event understanding. Detailed numbers can be found in the appendix.

\noindent \textbf{Qualitative Results and Discussion.}
Fig.~\ref{fig:finetune} presents a qualitative comparison of captions generated by Qwen2.5-VL-7B before and after fine-tuning, illustrating how fine-tuning enhances the model’s ability to capture causal relationships in traffic accident scenes. Before fine-tuning, the model produced broad scene-level descriptions, emphasizing general context such as weather, road environment, and visible agents. Although fluent and visually grounded, the captions failed to capture the key accident event, overlooked the vehicle–scooter interaction, and did not describe the cause or outcome of the collision.
After fine-tuning, the model generated more accurate and coherent accident narratives, it correctly identified the trigger event, namely the scooter rider entering the vehicle’s lane and making frontal contact with the car, and clearly explained the cause–effect chain, including the rider’s loss of balance, fall to the ground, and the vehicle stopping immediately afterward.
These results show that RoadSafe365 significantly improves the model’s causal reasoning, event ordering, and safety-centric understanding, highlighting the value of domain-specific fine-tuning for safety-critical multimodal reasoning.

\subsection{Cross-domain Generalization}
This section tests \name’s cross-domain generalization under two settings: (1) real-world$\rightarrow$real-world and (2) real-world$\rightarrow$synthetic.



\begin{table}[t]
\centering
\Large
\caption{Average VQA accuracy (\%) of Qwen2.5-VL-7B with different fine-tuning steps across four datasets in the VRU-Accident benchmark~\cite{kim2025vru}, namely 
\textbf{DADA}: DADA\_2000~\cite{fang2019dada}, \textbf{CAP}: CAP\_DATA~\cite{fang2022cognitive}, \textbf{DoTA}: DoTA~\cite{yao2022dota}, \textbf{MANU}: MANUAL\_DATA.}
\label{tab:vqa_steps_comparison}
\resizebox{\linewidth}{!}{%
\begin{tabular}{lccccc}
\toprule
\textbf{Model} & \textbf{Steps} & \textbf{DADA} & \textbf{CAP} & \textbf{DoTA} & \textbf{MANU} \\ 
\midrule

\multirow{4}{*}{\textbf{Qwen2.5-VL-7B}} 
& NF     & 59.27 & 62.72 & 57.50 & 59.79 \\

& 900    & 62.56 {\textcolor{ForestGreen}{↑3.29}} & 
            66.09 {\textcolor{ForestGreen}{↑3.37}} &
            59.67 {\textcolor{ForestGreen}{↑2.17}} &
            64.10 {\textcolor{ForestGreen}{↑4.31}} \\

& 1200   & 61.43 {\textcolor{ForestGreen}{↑2.16}} &
            67.02 {\textcolor{ForestGreen}{↑4.30}} &
            61.00 {\textcolor{ForestGreen}{↑3.50}} &
            63.29 {\textcolor{ForestGreen}{↑3.50}} \\

& 1500   & 61.14 {\textcolor{ForestGreen}{↑1.87}} &
            66.61 {\textcolor{ForestGreen}{↑3.89}} &
            60.50 {\textcolor{ForestGreen}{↑3.00}} &
            62.99 {\textcolor{ForestGreen}{↑3.20}} \\

\bottomrule
\end{tabular}
}
\end{table}

\noindent \textbf{Real-world $\rightarrow$ Real-world.}
We reuse the Qwen2.5-VL-7B model fine-tuned via VAU-R1 training and directly evaluate it on the real-world VRU-Accident benchmark~\cite{kim2025vru}. As shown in Table~\ref{tab:vqa_steps_comparison}, the fine-tuned models at $900, 1200$, and $1500$ steps consistently outperform the original Qwen2.5-VL-7B model, demonstrating that \name enhances the model’s understanding of real accidents—particularly in reasoning and agent interaction. These results further confirm that knowledge learned from RoadSafe365 could effectively transfer to a new real-world accident benchmark.

\begin{figure}[t]
    \centering
    \includegraphics[width=\columnwidth]{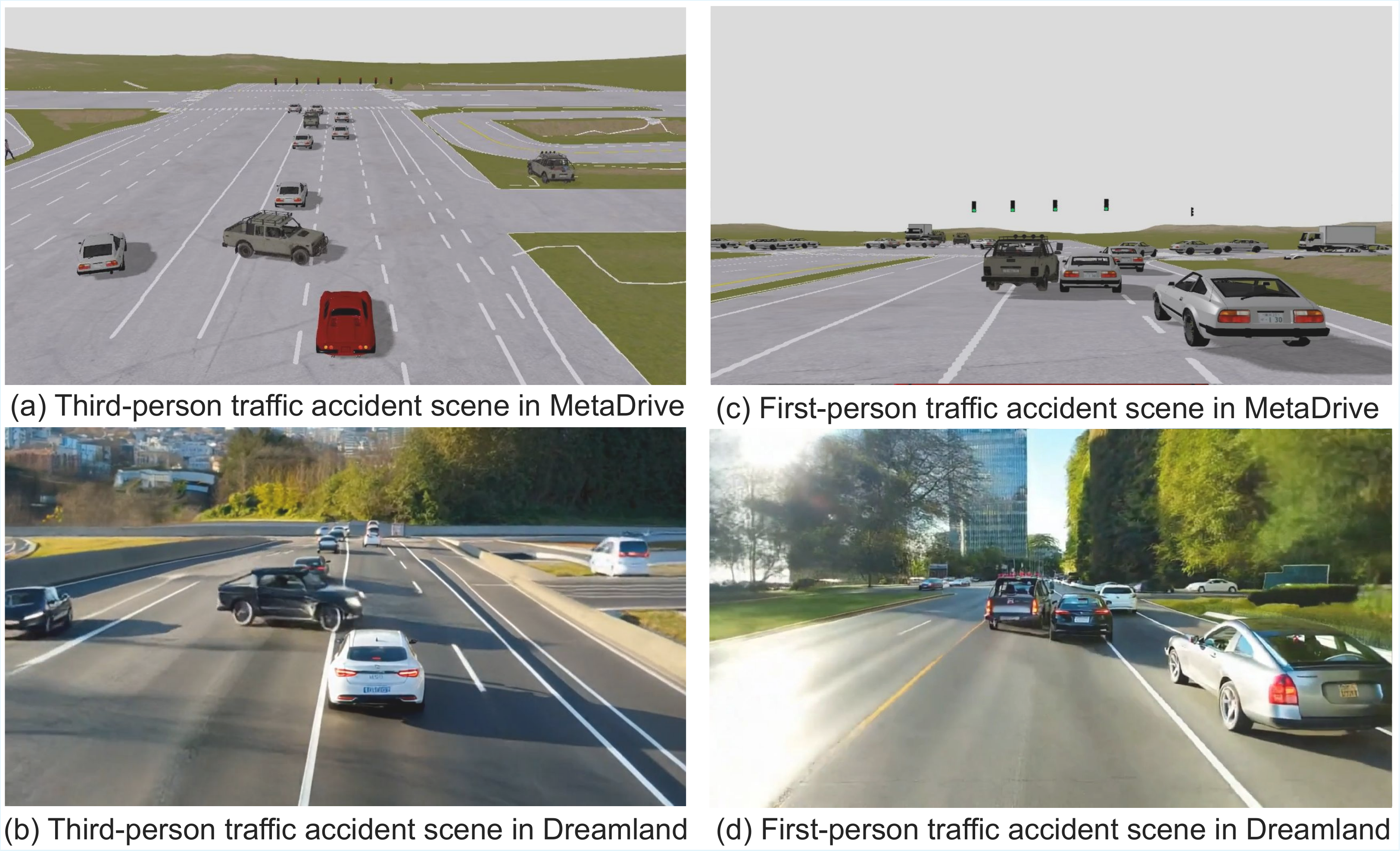}
    \caption{Sample visualizations of synthetic accident scenes from \name-Synthetic, created from MetaDrive~\cite{li2022metadrive} and Dreamland~\cite{mo2025dreamland}, in both first- and third-person views.}
    \label{fig:synthetic}
\end{figure}

\begin{figure}[t]
    \centering
    \includegraphics[width=\columnwidth]{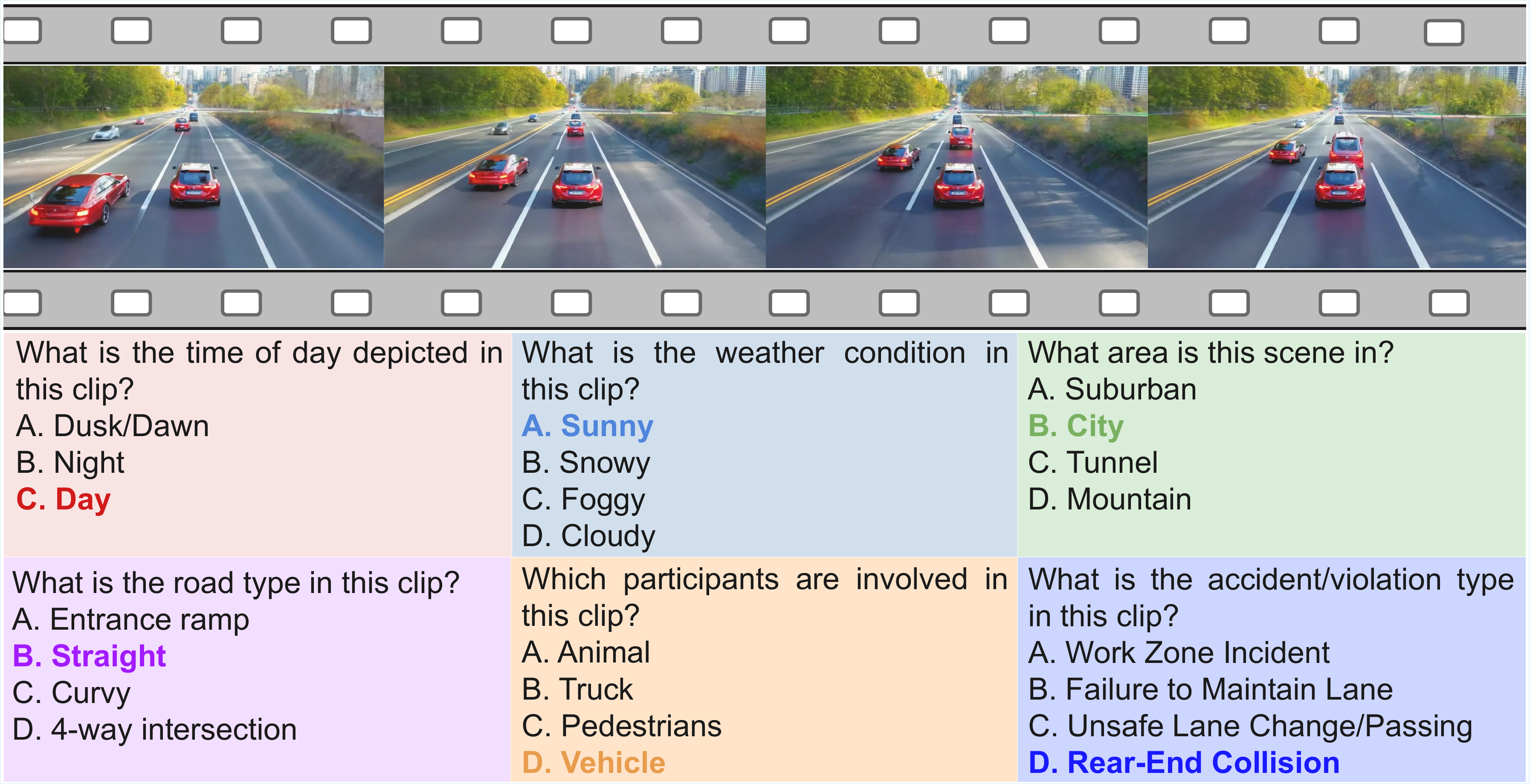}
    \caption{Visualization of annotations from Dreamland~\cite{mo2025dreamland} in both first-person and third-person views.}
    \label{fig:dreamland_vqa}
    \vspace{-1em}
\end{figure}

 \noindent \textbf{Real-world $\rightarrow$ Synthetic.} We further construct a realistic synthetic accident dataset via Adv-BMT method~\cite{liu2025adv}, into our \name-Synthetic, containing two simulation-based sources: (1) MetaDrive~\cite{li2022metadrive} clips and (2) Dreamland-rendered~\cite{mo2025dreamland} scenes, each offering first- and third-person views (Fig.~\ref{fig:synthetic}). Specifically, we generate $123$ third-person and $133$ first-person accident clips in MetaDrive, covering diverse road layouts, agent interactions, and accident types. Each scene is then rendered in Dreamland to produce high-fidelity videos with varied lighting conditions and visual styles. All clips are annotated using the RoadSafe365 pipeline, ensuring consistent annotations. Sample annotations could be found in Fig.~\ref{fig:dreamland_vqa}.



\begin{table}[t]
\centering
\scriptsize
\caption{Average VQA accuracy (\%) on synthetic first-person (FV) and third-person (TV) videos, rendered in MetaDrive and Dreamland under different training steps. 
\textbf{NF}: The original Qwen2.5-VL-7B model without reinforcement fine-tuning.}
\label{tab:vqa_metadrive_dreamland_tv}
\resizebox{\columnwidth}{!}{
\begin{tabular}{@{}lcccc@{}}
\toprule
\textbf{Dataset \& View} & \textbf{NF} & \textbf{900} & \textbf{1200} & \textbf{1500} \\ 
\midrule
MetaDrive--TV & 63.96 & 71.68\,\textcolor{ForestGreen}{↑7.72} & 73.98\,\textcolor{ForestGreen}{↑10.02} & 75.61\,\textcolor{ForestGreen}{↑11.65} \\
MetaDrive--FV & 64.41 & 76.19\,\textcolor{ForestGreen}{↑11.78} & 78.57\,\textcolor{ForestGreen}{↑14.16} & 78.32\,\textcolor{ForestGreen}{↑13.91} \\
Dreamland--TV & 72.95 & 77.44\,\textcolor{ForestGreen}{↑4.49} & 78.64\,\textcolor{ForestGreen}{↑5.69} & 79.16\,\textcolor{ForestGreen}{↑6.21} \\
Dreamland--FV & 67.47 & 74.89\,\textcolor{ForestGreen}{↑7.42} & 75.87\,\textcolor{ForestGreen}{↑8.40} & 76.19\,\textcolor{ForestGreen}{↑8.72} \\
\bottomrule
\end{tabular}
}
\end{table}

We again use the fine-tuned Qwen2.5-VL-7B model  and  evaluate it on the synthetic dataset. As shown in Table~\ref{tab:vqa_metadrive_dreamland_tv}, the model consistently surpasses the baseline across all four settings and model snapshots, achieving higher average VQA accuracy throughout.  These results highlight that \name fosters strong generalization to synthetic accident scenarios in various rendering styles and viewpoints. Detailed results are provided in the Appendix.

\definecolor{darkgreen}{RGB}{0,120,0} 

\definecolor{darkgreen}{RGB}{0,120,0}
\definecolor{darkred}{RGB}{180,0,0}
\newcommand{\gain}[1]{{\color{darkgreen}$\uparrow$#1}}
\newcommand{\loss}[1]{{\color{darkred}$\downarrow$#1}}

\section{Conclusion}
We have introduced
\name, a large-scale benchmark for real-world traffic safety understanding. It offers fine-grained annotations, a hierarchical taxonomy, and diverse QA tasks that jointly assess the perception, reasoning, and decision-making abilities of VLMs. Extensive experiments in both real-world and synthetic environments show that domain-specific training on RoadSafe365 substantially improves descriptive, reasoning, and generalization performance. RoadSafe365 bridges the gap between visual perception and reasoning in safety-critical environments, providing a unified testbed aligned with official safety standards. We envision this work contributing to the development of explainable, reliable, and human-aligned VLMs for intelligent transportation and autonomous driving.

{
    \small
    \bibliographystyle{ieeenat_fullname}
    \bibliography{main}
}

\clearpage

\newpage
\onecolumn
\addcontentsline{toc}{section}{Appendix} 
\renewcommand \thepart{} 
\renewcommand \partname{}
\part{\Large{\centerline{Appendix}}}
\parttoc

\newpage
\appendix


\label{Supplementary_Material}
This supplementary document provides detailed information about the RoadSafe365 benchmark. We organize the appendix as follows:

\begin{itemize}
\item \textbf{Annotation Pipeline Details.} We provide the exact prompts used for attribute labeling (VQA) and narrative captioning, detail the VQA distractor generation strategy, and offer justification for our annotation categories;
\item \textbf{Annotation Prompts.} We list the exact prompts used for VQA and dense captioning; 
\item \textbf{Dataset Statistics and Comparisons.} We offer additional data statistics, including detailed category distributions and visualizations, and provide a comprehensive comparison table against existing traffic-related datasets; 
\item \textbf{Fine-tuning Details.} We detail the setup for our fine-tuning experiments. We also include ablation studies on training steps; 
\item \textbf{Additional Qualitative Examples.} We show more visual examples of our annotations and qualitative comparisons of model outputs.
\end{itemize}

\section{Annotation Pipeline Details}
Our annotation pipeline combines large-model generation with a structured human verification process to ensure both scalability and high-quality labels.

\subsection{Stage 1: MLLMs-based Annotation}
We use two MLLMs for our initial annotation generation: Gemini 2.5 Pro for structured attributes (VQA) and GPT-4o for dense captions.

\noindent \textbf{Attribute Annotation.} For each video, we extract up to six keyframes. Our frame sampling selects three fixed timestamps (20\%, 50\%, 80\% of the video) and up to three frames with noticeable motion changes, detected by simple frame differencing. These frames are arranged into a 3×2 montage, which provides a compact visual summary of the incident.
The montage is then fed to Gemini 2.5 Pro with a structured prompt. The prompt enforces a fixed output format and a coarse-to-fine label structure that aligns with our taxonomy.

\noindent \textbf{Dense Caption Annotation.} For captions, we sample eight evenly spaced frames across the video. These frames are passed to GPT-4o with a targeted prompt. The model generates a 150–200 word paragraph in third-person past tense, describing only what is visually observable. The prompt instructs the model to describe the scene, the objects involved, and the visible event. 

\subsection{Stage 2: Human Verification and Correction}
All AI-generated attributes and captions are loaded into a lightweight web-based review tool. Reviewers access the tool locally in a browser.

\begin{figure*}[htb]
	\begin{minipage}[b]{1\textwidth}
		\centering		\includegraphics[width=0.98\textwidth]{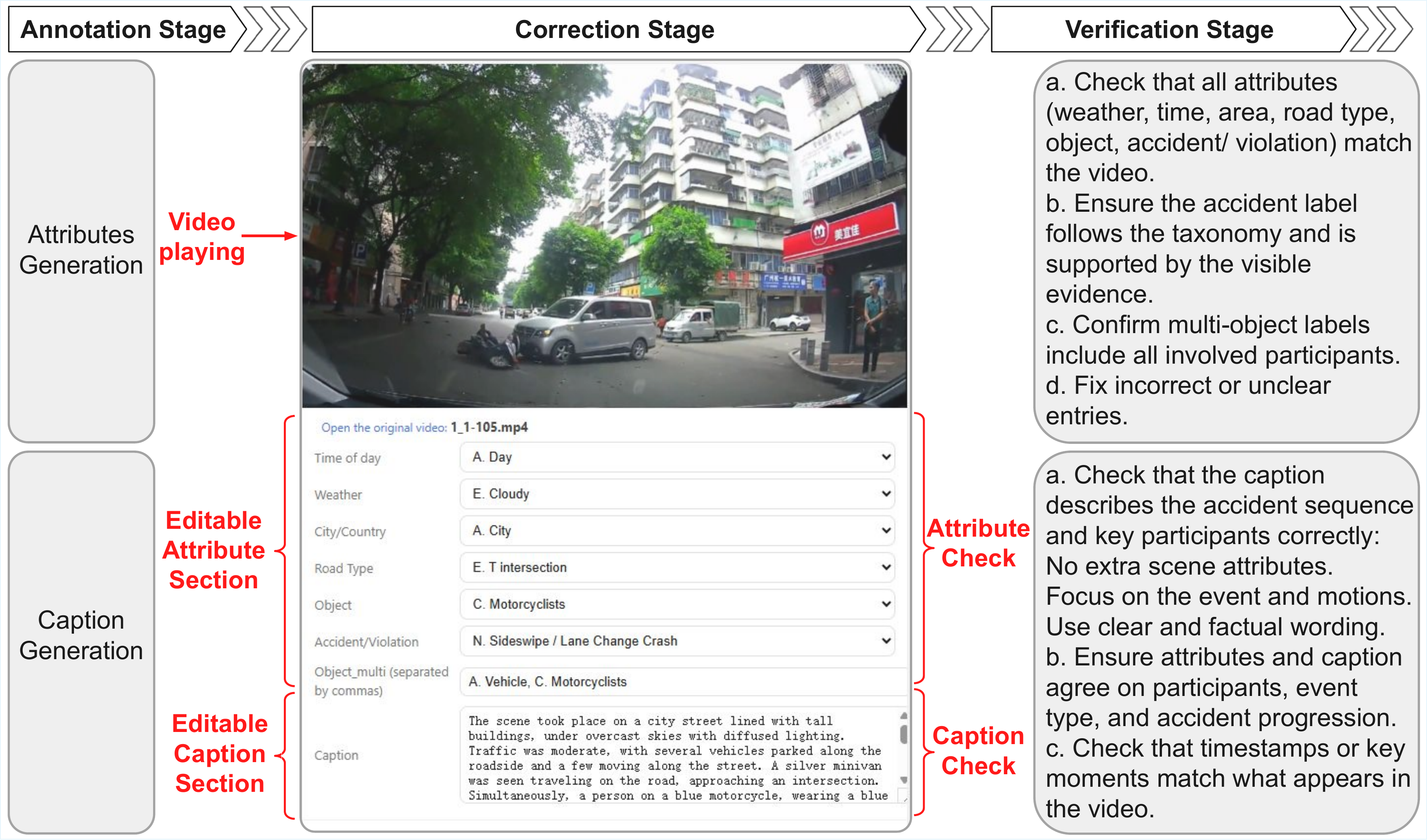}
	\end{minipage}
    \caption{\textbf{Overview of our MLLMs-assisted annotation and human verification interface.}The left side shows the generation stage, where attributes and a caption are produced by MLLMs. The right side shows the verification and modification stage.}
	\label{fig:correction} 
\end{figure*}

\noindent \textbf{Review Process.} As shown in Fig.~\ref{fig:correction}, the interface displays the video, the pre-filled attribute fields (as dropdown menus), and the caption (editable text box). Experts watch the video, check the MLLMs annotations for correctness, and directly adjust any incorrect fields or phrasing. All edits are saved automatically to the browser’s local storage. Once a batch is completed, all edits are collected into a single JSON file. 
Finally, a merge script processed this JSON file and updated the dataset by replacing the initial MLLMs-generated labels with the expert-verified ground truth.

\subsection{VQA Distractor Generation}
We generated challenging multiple-choice VQA pairs by creating three ``distractor'' options for each correct answer. Our pipeline does not sample randomly. Instead, it selects distractors from a pool of semantically similar, incorrect options. For example, for a ``Rear-End Collision'' ground truth, distractors are sampled from other collision types (e.g., ``Sideswipe'' ``Head-on'') rather than irrelevant options (e.g., ``Speeding''). This ensures all choices are plausible and require fine-grained reasoning.

\section{Annotation Prompts}
\noindent \textbf{VQA Generation Prompt.}
For each traffic-scene video, six fixed QA pairs are constructed to comprehensively evaluate a model’s understanding of environmental context, road configuration, participant involvement, and accident semantics.
Each question follows a multiple-choice format, where distractors are dynamically generated to ensure diversity across scenes.

The six prompts used for the VQA task are as follows:

\noindent \textbf{Time of Day:} “What is the time of day depicted in this clip?”

\noindent \textbf{Weather Condition:} “What is the weather condition in this clip?”

\noindent \textbf{Traffic Environment:} “What area is this scene in?”

\noindent \textbf{Road Configuration:} “What is the road type in this clip?”

\noindent \textbf{Participant Type:} “Which participants are involved in this clip?”

\noindent \textbf{Accident or Violation Type:} “What is the accident/violation type in this clip?”

\noindent \textbf{Dense Caption Promp.}
To generate detailed and coherent accident descriptions, we designed a structured instruction prompt that guides the model to focus strictly on visual evidence.
The prompt used for all models is as follows:

“You are an expert traffic incident annotator.
Describe only what is visually present in the video, in one coherent paragraph.
Write in third-person past tense, focusing on the visible sequence of actions.
Include scene context such as weather, lighting, road type, and surrounding traffic, followed by a description of participants (vehicles or pedestrians), their interactions, and the resulting event.
If a collision occurs, specify who contacted whom, the point of impact, and direct consequences.
If a violation occurs without a visible collision, describe the risky or illegal maneuver and the visible cues supporting it.”

\section{Dataset Statistics and Comparisons}
\noindent \textbf{Taxonomy coverage.} As shown in Table~\ref{tab:taxonomy_accuracy}, We present per-subtype accuracy on the Accident/Violation VQA task, grouped into five top-level families: Traffic Crashes, Safety Law Violations, Vulnerable Road User (VRU) Crashes, Traffic Violations, and Traffic Incidents. The model achieves relatively high accuracy on frequent categories such as Single-Vehicle (59.39\%), Rear-End (65.28\%), Pedestrian-Involved (71.13\%), Bicycle-Involved (80.95\%), Distracted Driving (85.00\%), and Red-Light / Stop-Sign (90.00\%). In contrast, rare and fine-grained types such as Micromobility-Involved (8.89\%), Failure to Maintain Lane (13.33\%), Illegal U-Turn / Turn (15.79\%), and Blocking Intersection (14.29\%) remain challenging.

\begin{table*}[t]
\centering
\caption{Comparison on \textbf{RoadSafe365 taxonomy-VQA}. Columns report accuracy (\%) for Accident/Violation.}
\scriptsize 
\label{tab:vqa_accidents_results}
\resizebox{\textwidth}{!}{%
\begin{tabular}{@{}llr|llr@{}}
\toprule
\multicolumn{1}{c}{\textbf{Category}} & \multicolumn{1}{c}{\textbf{Subtype (Left)}} & \multicolumn{1}{c|}{\textbf{Acc. (\%)}} & \multicolumn{1}{c}{\textbf{Category}} & \multicolumn{1}{c}{\textbf{Subtype (Right)}} & \multicolumn{1}{c}{\textbf{Acc. (\%)}} \\ \midrule
Traffic Crashes & Single-Vehicle & 59.39 & Traffic Violations & Unsafe Lane Change / Passing & 35.59 \\
 & Multi-Vehicle & 26.23 &  & Failure to Yield & 60.20 \\
 & Rear-End & 65.28 &  & Reckless / Aggressive Driving & 27.27 \\
 & Sideswipe & 41.06 &  & Unsafe Merge / Entry & 41.82 \\
 & Angle / T-Bone & 37.98 &  & Speeding & 50.00 \\
 & Head-On & 22.67 &  & Tailgating & 87.50 \\
 & Intersection-Related & 52.78 &  & Failure to Maintain Lane & 13.33 \\
 & Roadway Departure & 58.70 &  & Illegal U-Turn / Turn & 15.79 \\
Safety Law Violations & Distracted Driving & 85.00 &  & Red-Light / Stop-Sign & 90.00 \\
 & Drowsy / Fatigued & 30.29 &  & Driving Wrong Way & 25.00 \\
 & Impaired Driving & 35.29 &  & Blocking Intersection & 14.29 \\
Vulnerable Road User Crashes & Pedestrian-Involved & 71.13 & Traffic Incidents & Obstruction / Road Hazard & 27.27 \\
 & Bicycle-Involved & 80.95 &  & Work-Zone Incident & 27.31 \\
 & Animal-Involved & 40.00 &  & General Incident & 30.42 \\
 & Micromobility-Involved & 8.89 &  & & \\ 
\bottomrule
\end{tabular}
}
\label{tab:taxonomy_accuracy}
\end{table*}

\noindent \textbf{Comparison with existing benchmarks.} As shown in Table~\ref{tab:datasets_comparison}, We compare RoadSafe365 with prior traffic datasets along view type, number of clips, and annotation modalities. RoadSafe365 jointly supports accident taxonomy, scene attributes, VQA, and dense captioning on both dashcam and surveillance videos, complementing recent benchmarks such as VRU-Accident, RoadSocial, and TAU-106K.

\begin{table*}[t]
\centering
\caption{Comparison of RoadSafe365 with existing traffic incident datasets. Dash.: Dashcam View, Surv.: Surveillance View, VQA: Video Question Answering, Attr.: Scene Attributes, Acc.-Tax.: Accident Taxonomy, DC: Dense Caption, R/S: Real or Synthetic videos. 
}
\tiny 
\resizebox{\textwidth}{!}{
\begin{tabular}{@{}lccccccccccccc@{}}
\toprule
\multicolumn{1}{c}{\textbf{Dataset}} & \textbf{Year} & \textbf{Data Source} & \textbf{Video Length} & \textbf{Label Taxonomy} & \textbf{Unique Features} & \textbf{View} & \textbf{\#Clips} & \textbf{Acc.-Tax.} & \textbf{Attr.} & \textbf{VQA} & \textbf{DC} & \textbf{R/S} \\ \midrule
A3D~\cite{yao2019unsupervised} & 2019 & Web (Dash) & 8.5 & Binary (Anomaly) & Anomaly Detection & Dash. & 1,500 & – & \checkmark & – & – & R \\
DADA-2000~\cite{fang2019dada} & 2019 & Web (Dash) & 11.0 & Accident Types & Driver Attention & Dash. & 2,000 & – & \checkmark & – & – & R \\
CTA~\cite{you2020traffic} & 2020 & Web (Dash) & 4.7 & Causal Labels & Causality Reasoning & Dash./Surv. & 1,935 & – & – & – & – & R \\
SUTD-TrafficQA~\cite{xu2021sutd} & 2021 & Surveillance & 13.6 & Event-based & Complex Logic QA & Surv. & 10,080 & – & – & \checkmark & – & R \\
DoTA~\cite{yao2022dota} & 2022 & Web (Dash) & 15.6 & Anomaly Types & Unsupervised Det. & Dash. & 5,586 & \checkmark & \checkmark & – & – & R \\
MM-AU~\cite{fang2022cognitive} & 2022 & Web (Dash) & 5.0 & Anticipation Labels & Accident Anticipation & Dash. & 11,727 & – & \checkmark & – & – & R \\
ROL~\cite{karim2023attention} & 2023 & Web (Dash) & 4.0 & Risk Labels & Risky Object Loc. & Dash. & 1,000 & – & \checkmark & – & – & R \\
CTAD~\cite{luo2023simulation} & 2023 & Simulation & 8.0 & Binary & Simulation-based & Surv. & 1,100 & – & \checkmark & – & – & S \\
TUMTraffic-VideoQA~\cite{zhou2025tumtraffic} & 2025 & Surveillance & 10.0 & Action/Event & Spatio-temporal QA & Surv. & 1,000 & – & – & \checkmark & – & R \\
TUMTraf-A~\cite{zimmer2025towards} & 2025 & Multi-sensor & -- & Frame-level & 3D Perception & Surv. & 48 & – & – & – & – & R \\
VRU-Accident~\cite{kim2025vru} & 2025 & Web (Dash) & 9.5 & VRU Types & VRU Focus & Dash. & 1,000 & \checkmark & \checkmark & \checkmark & \checkmark & R \\
RoadSocial~\cite{parikh2025roadsocial} & 2025 & Social (Dash) & 15.0 & Social Events & Social Narratives & Dash./Surv. & 13,200 & – & \checkmark & \checkmark & – & R \\
TAD-106K~\cite{zhou2025tau} & 2025 & Web (Dash) & 10.3 & Coarse Accident Labels & Large-scale accident collection & Dash./Surv. & 51,544 & \checkmark & \checkmark & \checkmark & \checkmark & R \\
\textbf{RoadSafe365 (Ours)} & 2025 & \textbf{Social (Mix)} & \textbf{11.2} & \textbf{NSC/ANSI Aligned} & \textbf{Standard-aligned Safety} & Dash./Surv. & 36,196 & \checkmark & \checkmark & \checkmark & \checkmark & R\&S \\ \bottomrule
\end{tabular}
}
\label{tab:datasets_comparison}
\end{table*}

\definecolor{darkgreen}{RGB}{0,120,0} 

\begin{table*}[t]
\centering
\caption{VQA performance of Qwen2.5-VL-7B fine-tuned on RoadSafe365. ``Change'' shows the absolute improvement over the non-finetuned model (NF). \textbf{Acc./Vio.}: Accident/Violation, \textbf{Obj.}: Object.}
\begin{tabular}{@{}lcccccccccc@{}}
\toprule
\textbf{Model} & \textbf{Steps} & \textbf{Weather} & \textbf{Acc./Vio.} & \textbf{Time} & \textbf{Area} & \textbf{Road} & \textbf{Obj.} & \textbf{Avg.} & \textbf{Change} \\ \midrule
\multirow{8}{*}{\textbf{Qwen2.5-VL-7B}} 
& NF    & 66.73 & 38.04 & 94.77 & 68.98 & 65.17 & 35.70 & 61.57 & -- \\ \cmidrule(l){2-10}
& 300   & 78.29 & 48.14 & 93.57 & 78.57 & 73.86 & 61.14 & 72.26 & {\color{darkgreen}{$\uparrow$10.69}} \\
& 500   & 74.14 & 60.86 & 90.71 & 88.29 & 75.57 & 70.29 & 76.64 & {\color{darkgreen}{$\uparrow$15.07}} \\
& 600   & 77.00 & 62.43 & 98.14 & 88.43 & 73.14 & 74.57 & 78.95 & {\color{darkgreen}{$\uparrow$17.38}} \\
& 900   & 79.29 & 61.57 & 98.14 & 92.00 & 75.00 & 84.71 & 81.79 & {\color{darkgreen}{$\uparrow$20.22}} \\
& 1000  & 78.14 & 54.29 & 98.14 & 91.57 & 73.71 & 84.71 & 80.10 & {\color{darkgreen}{$\uparrow$18.53}} \\
& 1200  & 79.71 & 66.00 & 98.43 & 91.43 & 71.43 & 83.14 & 81.69 & {\color{darkgreen}{$\uparrow$20.12}} \\
& 1500  & 79.86 & 67.29 & 98.14 & 92.00 & 67.86 & 82.14 & 81.21 & {\color{darkgreen}{$\uparrow$19.64}} \\ 
\bottomrule
\end{tabular}%
\label{tab:vqa_finetune_breakdown}
\end{table*}

\section{Fine-tuning Details}
\subsection{Baseline Models}
We fine-tune Qwen2.5-VL-7B as our main open-source baseline. We use the same optimizer, learning rate schedule, and loss functions as in the main paper. We include the full VQA results in Table~\ref{tab:vqa_finetune_breakdown}, which expands the summary reported in the main paper.

\subsection{Evaluation Metrics}
For the VQA task, we use accuracy as the main metric. For the dense captioning task, we report SPICE, METEOR, COMET, and ROUGE-L, since they capture meaning better than surface word matching. 
SPICE checks the meaning of a caption by breaking it into objects, attributes, and relations, and comparing them with the ground truth. 
METEOR measures word alignment with the reference and considers stemming and synonyms.
COMET uses a pre-trained language model to assess how close the caption’s meaning is to the reference.
ROUGE-L measures the longest common subsequence between the prediction and the reference, focusing on recall and sentence structure.

\subsection{Cross-dataset Generalization}
To assess cross-dataset generalization, we evaluate the non-finetuned (NF) model and the 900-step model on VRU-Accident dataset which contains DADA\_2000, CAP\_DATA, DoTA, and a manually collected set of videos. As shown in Table~\ref{tab:vqa_breakdown_nf_vs_900}, we report accuracy on weather, location, road type, accident type, accident reason, and prevention method, following the same VQA formulation as in \name.

Across all datasets, the fine-tuned model shows consistent gains in scene understanding (weather and location) and moderate improvements in high-level reasoning (accident reason and prevention). Accuracy on accident-type questions remains relatively stable.

\definecolor{darkgreen}{RGB}{0,120,0}
\definecolor{darkred}{RGB}{180,0,0}

\begin{table*}[t]
\centering
\caption{Breakdown of VQA performance comparison between the non-finetuned (NF) Qwen2.5-VL-7B and the model fine-tuned for 900 steps across four datasets in the VRU-Accident benchmark~\cite{kim2025vru}. All metrics are accuracy (\%). "Change" indicates the absolute percentage point difference. \textbf{Road}: Road Type, \textbf{Acc.}: Accident Type, \textbf{Acc.R}: Accident Reason, \textbf{Prev.}: Prevention Method.}
\begin{tabular}{@{}llrrrrrrr@{}}
\toprule
\textbf{Dataset} & \textbf{Model} & \textbf{Weather} & \textbf{Location} & \textbf{Road} & \textbf{Acc.} & \textbf{Acc.R} & \textbf{Prev.} & \textbf{Avg.} \\ 
\midrule
\multirow{3}{*}{\textbf{DADA\_2000}~\cite{fang2019dada}} 
& NF & 61.43 & 82.51 & 65.92 & 59.19 & 34.98 & 51.57 & 59.27 \\
& 900 & 68.61 & 90.58 & 62.78 & 60.99 & 38.12 & 54.26 & 62.56 \\
& Change & \gain{7.18} & \gain{8.07} & \loss{3.14} & \gain{1.80} & \gain{3.14} & \gain{2.69} & \gain{3.29} \\
\midrule
\multirow{3}{*}{\textbf{CAP\_DATA}~\cite{fang2022cognitive}} 
& NF & 62.37 & 77.00 & 74.91 & 56.79 & 46.34 & 58.89 & 62.72 \\
& 900 & 73.87 & 86.41 & 70.38 & 55.75 & 51.57 & 58.54 & 66.09 \\
& Change & \gain{11.50} & \gain{9.41} & \loss{4.53} & \loss{1.04} & \gain{5.23} & \loss{0.35} & \gain{3.37} \\
\midrule
\multirow{3}{*}{\textbf{DoTA}~\cite{yao2022dota}} 
& NF & 59.00 & 82.00 & 71.00 & 52.00 & 26.00 & 55.00 & 57.50 \\
& 900 & 68.00 & 89.00 & 72.00 & 48.00 & 22.00 & 59.00 & 59.67 \\
& Change & \gain{9.00} & \gain{7.00} & \gain{1.00} & \loss{4.00} & \loss{4.00} & \gain{4.00} & \gain{2.17} \\
\midrule
\multirow{3}{*}{\textbf{MANUAL\_DATA}~\cite{kim2025vru}} 
& NF & 63.85 & 74.10 & 72.31 & 56.67 & 40.77 & 51.03 & 59.79 \\
& 900 & 67.44 & 90.77 & 70.26 & 55.64 & 42.05 & 58.46 & 64.10 \\
& Change & \gain{3.59} & \gain{16.67} & \loss{2.05} & \loss{1.03} & \gain{1.28} & \gain{7.43} & \gain{4.31} \\
\bottomrule
\end{tabular}%
\label{tab:vqa_breakdown_nf_vs_900}
\end{table*}

\subsection{Synthetic-scene Evaluation}
As shown in Table~\ref{tab:synthetic_third_view_vqa} and Table~\ref{tab:synthetic_vqa_avg}, we report per-attribute accuracy on synthetic third-person-view accident videos and summarize average accuracy on first-person (FV) and third-person (TV) views from MetaDrive and DreamLand. In addition to the results reported in the main paper, we further provide per-attribute accuracy on synthetic third-person-view accident videos and include extended comparisons at 600 and 1000 training steps.

\subsection{Data-scale Ablation on RoadSafe365}
To examine how the amount of training data influences performance, we run a data-scale ablation on RoadSafe365. We fine-tune Qwen2.5-VL-7B using three subsets of the training split: 1,500, 3,000, and 6,000 videos, while keeping the validation set fixed. Table~\ref{tab:Data-scale_Ablation} shows that using 3k videos already provides most of the gains at all three training steps (300, 600, 900). Scaling the data to 6k videos brings smaller but steady improvements. In contrast, using only 1.5k videos leads to clear drops in accuracy, especially at the early steps.

\begin{table}[t]
\centering
\scriptsize
\caption{
Comparison of VQA accuracy (\%) under different training data scales 
(1.5k, 3k, 6k videos) and fine-tuning steps (300, 600, 900). 
Values for 1.5k and 3k are placeholders. ``Avg.'' is the mean VQA accuracy 
across six categories.
}
\label{tab:data_scale_vs_steps}
\resizebox{0.4\columnwidth}{!}{
\begin{tabular}{@{}lccc@{}}
\toprule
\textbf{Training Scale} & \textbf{300 steps} & \textbf{600 steps} & \textbf{900 steps} \\ 
\midrule
\textbf{1500} & 70.60 & 73.43 & 74.67 \\[1mm]
\textbf{3000} & 71.05 & 76.07 & 79.21 \\[1mm]
\textbf{6000} 
& 72.26 
& 78.95 
& 81.79 \\ 
\bottomrule
\end{tabular}
}
\label{tab:Data-scale_Ablation}
\end{table}

To further assess the robustness of the observed trends, as shown in Table~\ref{tab:ft_12k_steps}, we re-ran fine-tuning using a larger training subset of 12k clips. While the average accuracy is modestly higher than in the 6k setting, the learning dynamics and relative performance trends remain consistent. This suggests that the observed gains are stable and not attributable to small-sample effects or training noise.

\begin{table}[htb]
\centering
\caption{Average VQA accuracy (\%) of Qwen models fine-tuned on 6k and 12k RoadSafe365 clips under different training steps.}
\vspace{-0.15cm}
\footnotesize
\resizebox{0.5\columnwidth}{!}{%
\begin{tabular}{c|ccccc}
\toprule
\textbf{Train Size} & \textbf{300} & \textbf{500} & \textbf{600} & \textbf{900} & \textbf{1000} \\
\midrule
6k clips (original) & 72.26 & 76.64 & 78.95 & 81.79 & 80.10 \\
12k clips & 72.40 & 75.40 & 78.60 & 80.55 & 81.12 \\
\bottomrule
\end{tabular}
}
\label{tab:ft_12k_steps}
\end{table}

\noindent \textbf{Evaluation protocol and temporal evidence.}
We adopt a 700-clip, keyframe-based test set to balance temporal reasoning fidelity and inference cost. A frame ablation study (Table~\ref{tab:frame_ablation}, 1K-step training) shows consistent accuracy gains with additional frames. Single-frame sampling is temporally sensitive: the first frame performs worst, while the middle frame performs best. This aligns with the temporal structure of many events, which tend to occur near the center of clips rather than at their boundaries. Nevertheless, they all remain substantially inferior to multi-frame inputs, demonstrating that static frames alone are insufficient and that temporal context is critical. We select 8 frames as an effective accuracy--efficiency trade-off. The test set is explicitly constructed to ensure sufficient coverage of all Level-2 subcategories, including rare ones, reducing bias toward frequent events.


\subsection{Backbone Robustness with Qwen3-VL}
To examine whether our conclusions depend on a specific vision-language backbone, we additionally evaluate a more recent model, Qwen3-VL-4B, using the same fine-tuning pipeline and training data. Due to computational constraints, we report average VQA accuracy under different fine-tuning steps.

As shown in Table~\ref{tab:qwen3_avg}, Qwen3-VL exhibits consistent performance trends compared to Qwen2.5-VL across training steps. While absolute accuracy differs due to model capacity, the relative improvements with additional fine-tuning remain similar. This suggests that our observations are robust across model generations and not tied to a specific backbone.

\section{Additional Qualitative Examples}
We provide additional qualitative examples to illustrate the diversity and coverage of our benchmark. For each of the five top-level families in our taxonomy, we select representative subcategories and show their corresponding video clips together with the finalized annotations.

\clearpage
\newpage

\begin{table}
\centering
\caption{Frame ablation on the 700-clip test set. \textbf{1$_{\text{first}}$} and \textbf{1$_{\text{last}}$} are the first or last frame, while \textbf{1} is the middle frame.}
\vspace{-0.15cm}
\footnotesize
\resizebox{0.7\columnwidth}{!}{%
\begin{tabular}{c|cccccccc}
\toprule
\textbf{\#Frames} & \textbf{1$_{\text{first}}$} & \textbf{1$_{\text{last}}$} & \textbf{1} & \textbf{2} & \textbf{4} & \textbf{8} & \textbf{16} & \textbf{32} \\
\midrule
Acc./Vio. & 22.86 & 41.14 & 49.34 & 52.14 & 53.02 & 54.29 & 55.29 & 56.86 \\
Remaining Attrs. Avg. & 65.46 & 79.46 & 79.62 & 81.00 & 83.46 & 85.26 & 87.37 & 87.23 \\
\midrule
Avg. Accuracy & 58.36 & 73.07 & 74.57 & 76.19 & 78.39 & 80.10 & 82.02 & 82.17 \\
\bottomrule
\end{tabular}%
}
\label{tab:frame_ablation}
\vspace{-.5em}
\end{table}

\begin{table*}[t]
\centering
\caption{VQA accuracy (\%) on synthetic third-person view traffic accident videos across different training steps for MetaDrive. \textbf{NF}: No fine-tune.}
\begin{tabular}{@{}ccccccccc@{}}
\toprule
\textbf{Model} & \textbf{Steps} & \textbf{City} & \textbf{Object} & \textbf{Acc.} & \textbf{Road} & \textbf{Time} & \textbf{Weather} & \textbf{Avg.} \\ 
\midrule
\multirow{6}{*}{Qwen2.5-VL-7B} 
 & NF   & 69.92 & 44.72  & 53.66 & 67.48 & 100.00 & 47.97 & 63.96 \\
 \cmidrule(l){2-9} 
 & 600  & 68.29 & 65.04  & 39.02 & 65.85 & 100.00 & 59.35 & 66.26 \\
 & 900  & 71.54 & 100.00 & 36.59 & 66.67 & 100.00 & 55.28 & 71.68 \\
 & 1000 & 70.73 & 99.19  & 39.02 & 67.48 & 100.00 & 71.54 & 74.66 \\
 & 1200 & 72.36 & 99.19  & 37.40 & 63.41 & 100.00 & 71.54 & 73.98 \\
 & 1500 & 69.92 & 100.00 & 41.46 & 63.41 & 100.00 & 78.86 & 75.61 \\
\bottomrule
\end{tabular}%
\label{tab:synthetic_third_view_vqa}
\end{table*}

\begin{table*}[t]
\centering
\caption{Average VQA accuracy (\%) across first-person (FV) and third-person (TV) traffic accident videos for MetaDrive and DreamLand under different training steps. \textbf{NF}: original Qwen2.5-VL-7B model without reinforcement fine-tuning.}
\begin{tabular}{@{}lccccccc@{}}
\toprule

\textbf{Dataset \& View} & \textbf{NF} & \textbf{600} & \textbf{900} & \textbf{1000} & \textbf{1200} & \textbf{1500} \\ 
\midrule 
MetaDrive–TV & 63.96 & 66.26 & 71.68 & 74.66 & 73.98 & 75.61 \\
MetaDrive–FV & 64.41 & 72.93 & 76.19 & 80.20 & 78.57 & 78.32 \\ 
DreamLand–TV & 72.95 & 74.40 & 77.44 & 80.57 & 78.64 & 79.16 \\
DreamLand–FV & 67.47 & 67.95 & 74.04 & 76.92 & 75.00 & 75.16 \\ 
\bottomrule
\end{tabular}%
\label{tab:synthetic_vqa_avg}
\vspace{-1em}
\end{table*}

\vspace{-0.25cm}
\begin{table}[htb]
\centering
\caption{Average VQA accuracy (\%) of Qwen models under different fine-tuning steps using the same training pipeline.}
\vspace{-0.2cm}
\footnotesize
\resizebox{0.45\columnwidth}{!}{%
\begin{tabular}{c|ccc}
\toprule
\textbf{Backbone} & \textbf{500 steps} & \textbf{1000 steps} & \textbf{1500 steps} \\
\midrule
Qwen2.5-VL-7B & 76.64 & 80.10 & 81.21 \\
Qwen3-VL-4B & 81.55 & 83.81 & 83.36 \\
\bottomrule
\end{tabular}
}
\vspace{-1em}
\label{tab:qwen3_avg}
\end{table}

\clearpage
\newpage
\begin{figure*}[htb]
	\begin{minipage}[b]{1\textwidth}
		\centering		\includegraphics[width=0.98\textwidth]{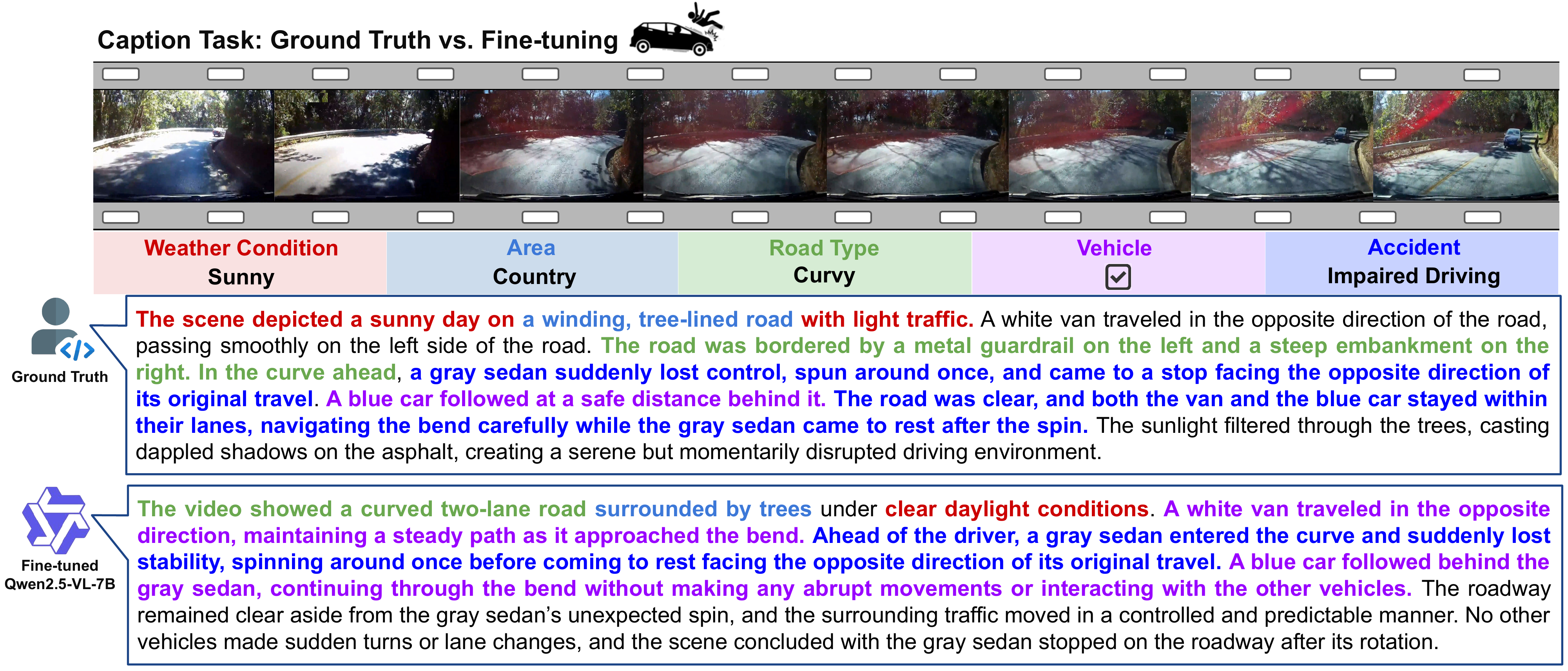}
	\end{minipage}
    \caption{Examples of accident cases across the five major categories in RoadSafe365.}
	\label{fig:vqa} 
\end{figure*}

\begin{figure*}[htb]
	\begin{minipage}[b]{1\textwidth}
		\centering		\includegraphics[width=0.98\textwidth]{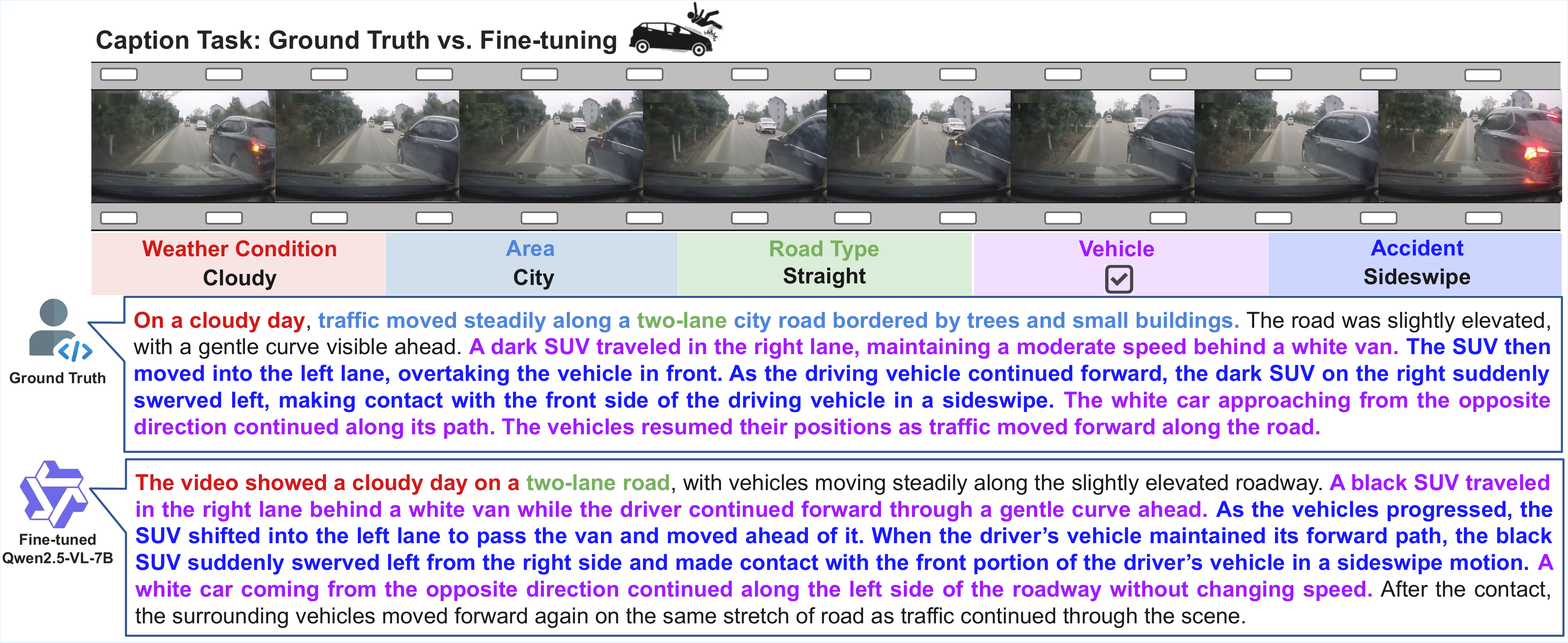}
	\end{minipage}
    \caption{Examples of accident cases across the five major categories in RoadSafe365.}
	\label{fig:vqa} 
\end{figure*}

\begin{figure*}[htb]
	\begin{minipage}[b]{1\textwidth}
		\centering		\includegraphics[width=0.98\textwidth]{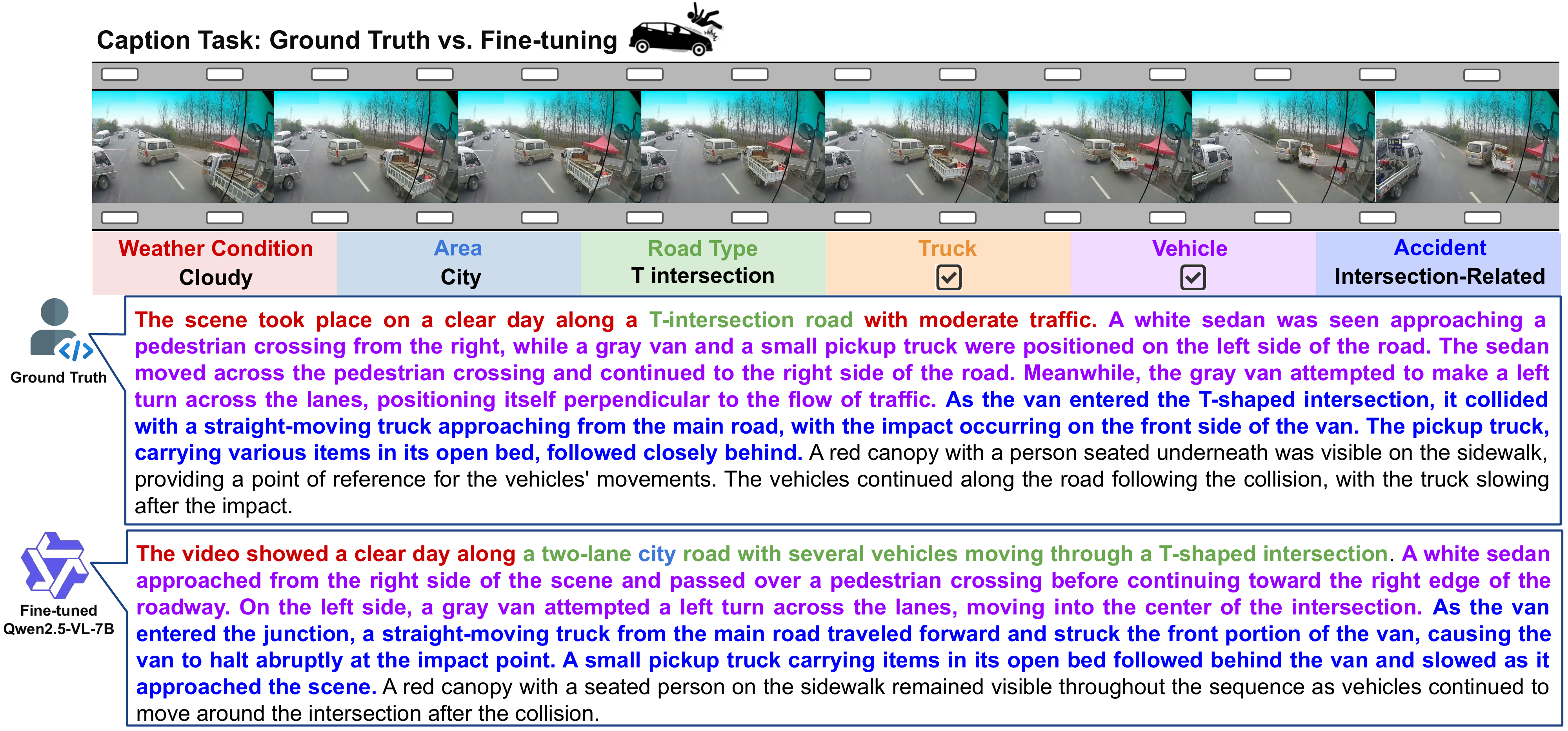}
	\end{minipage}
    \caption{Examples of accident cases across the five major categories in RoadSafe365.}
	\label{fig:vqa} 
\end{figure*}

\begin{figure*}[htb]
	\begin{minipage}[b]{1\textwidth}
		\centering		\includegraphics[width=0.98\textwidth]{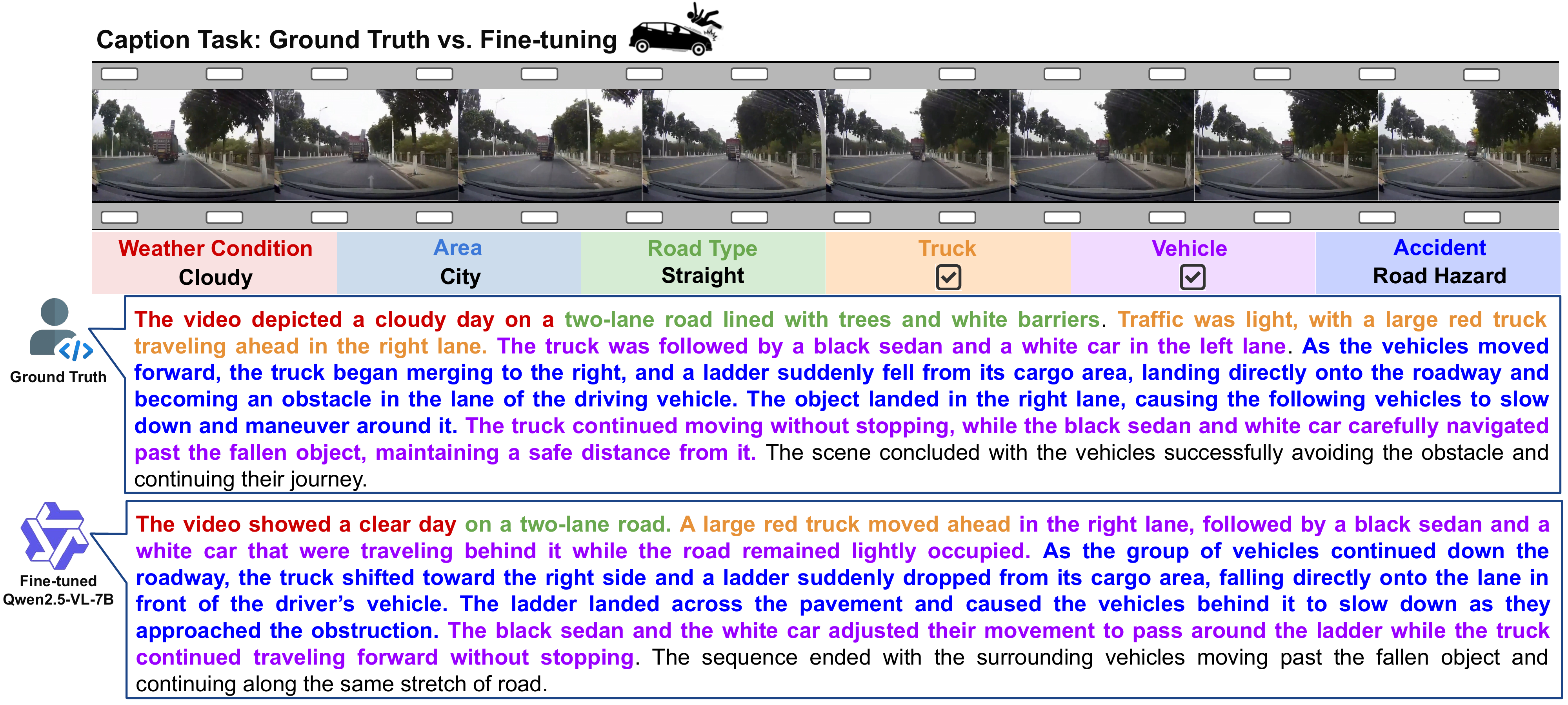}
	\end{minipage}
    \caption{Examples of accident cases across the five major categories in RoadSafe365.}
	\label{fig:vqa} 
\end{figure*}

\begin{figure*}[htb]
	\begin{minipage}[b]{1\textwidth}
		\centering		\includegraphics[width=0.98\textwidth]{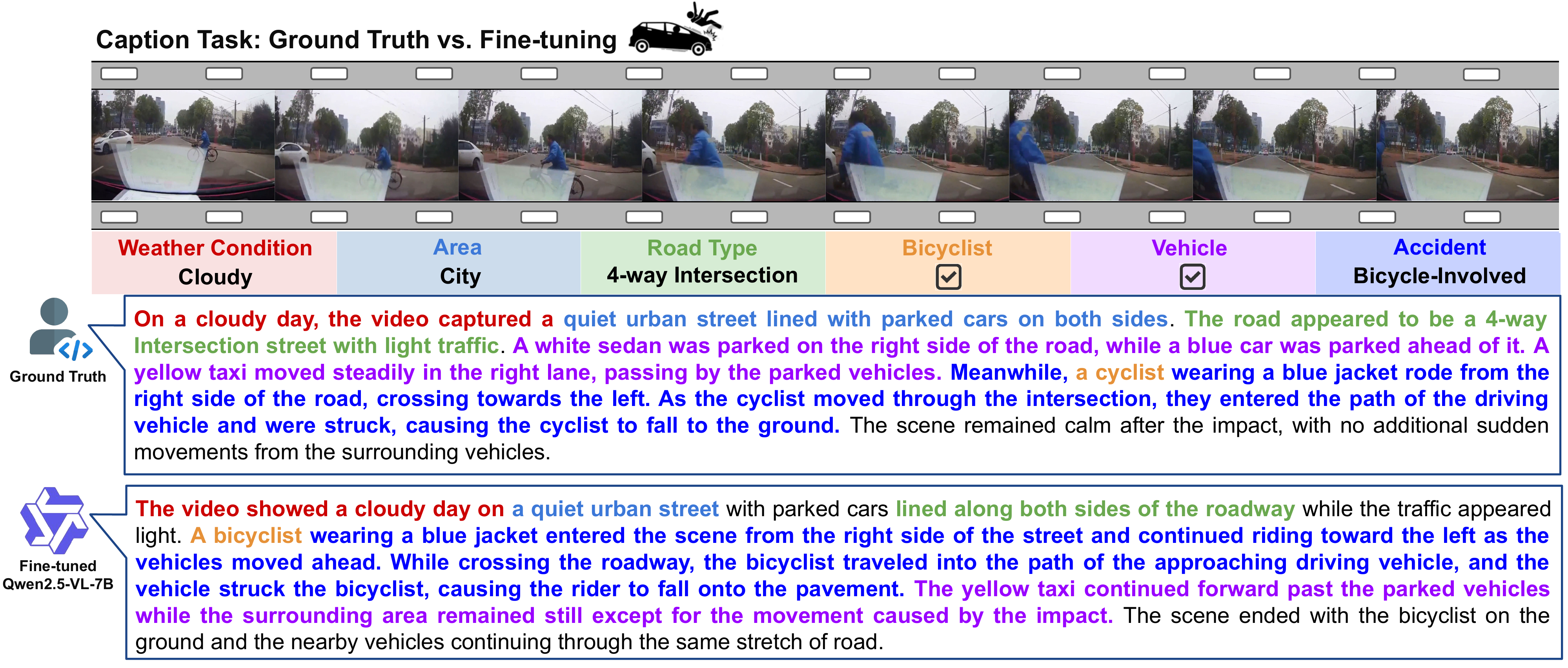}
	\end{minipage}
    \caption{Examples of accident cases across the five major categories in RoadSafe365.}
	\label{fig:vqa} 
\end{figure*}

\begin{figure*}[htb]
	\begin{minipage}[b]{1\textwidth}
		\centering		\includegraphics[width=0.98\textwidth]{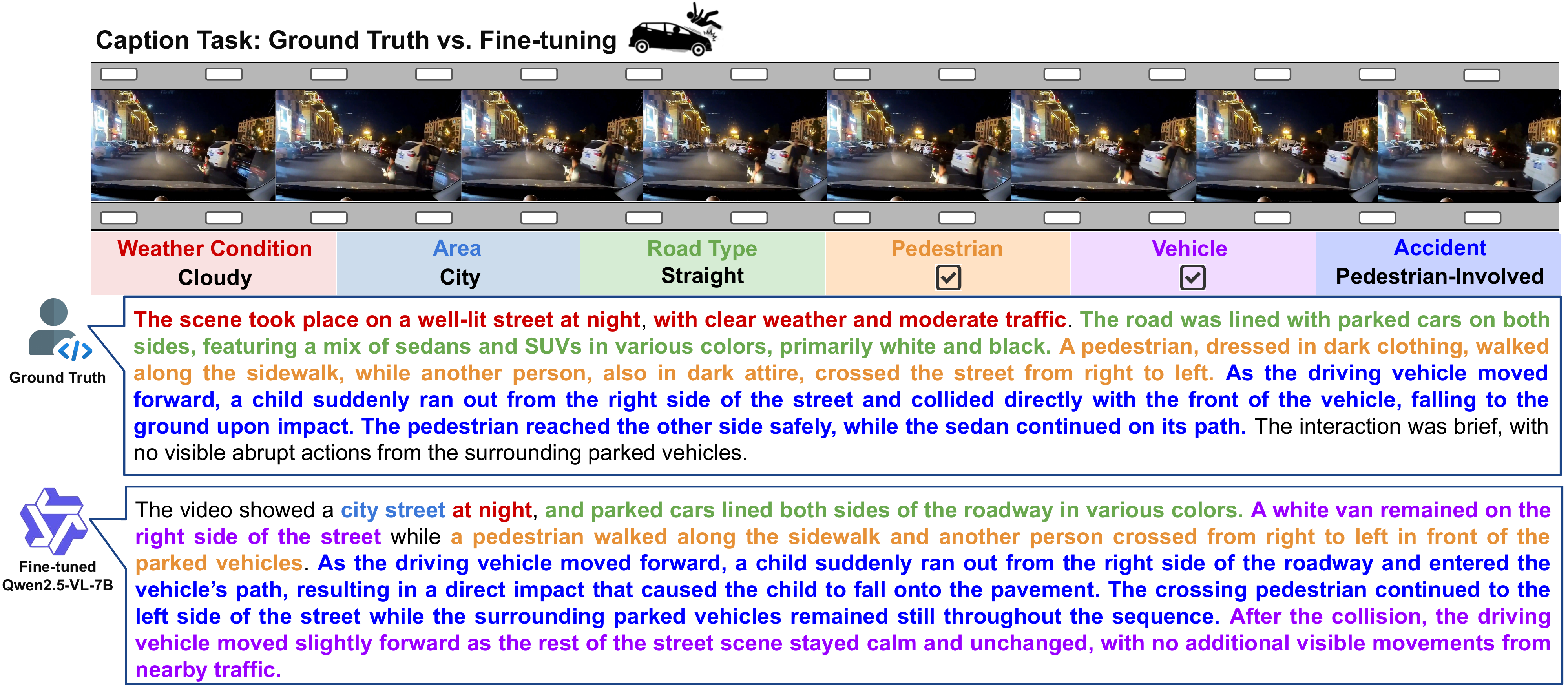}
	\end{minipage}
    \caption{Examples of accident cases across the five major categories in RoadSafe365.}
	\label{fig:vqa} 
\end{figure*}

\begin{figure*}[htb]
	\begin{minipage}[b]{1\textwidth}
		\centering		\includegraphics[width=0.98\textwidth]{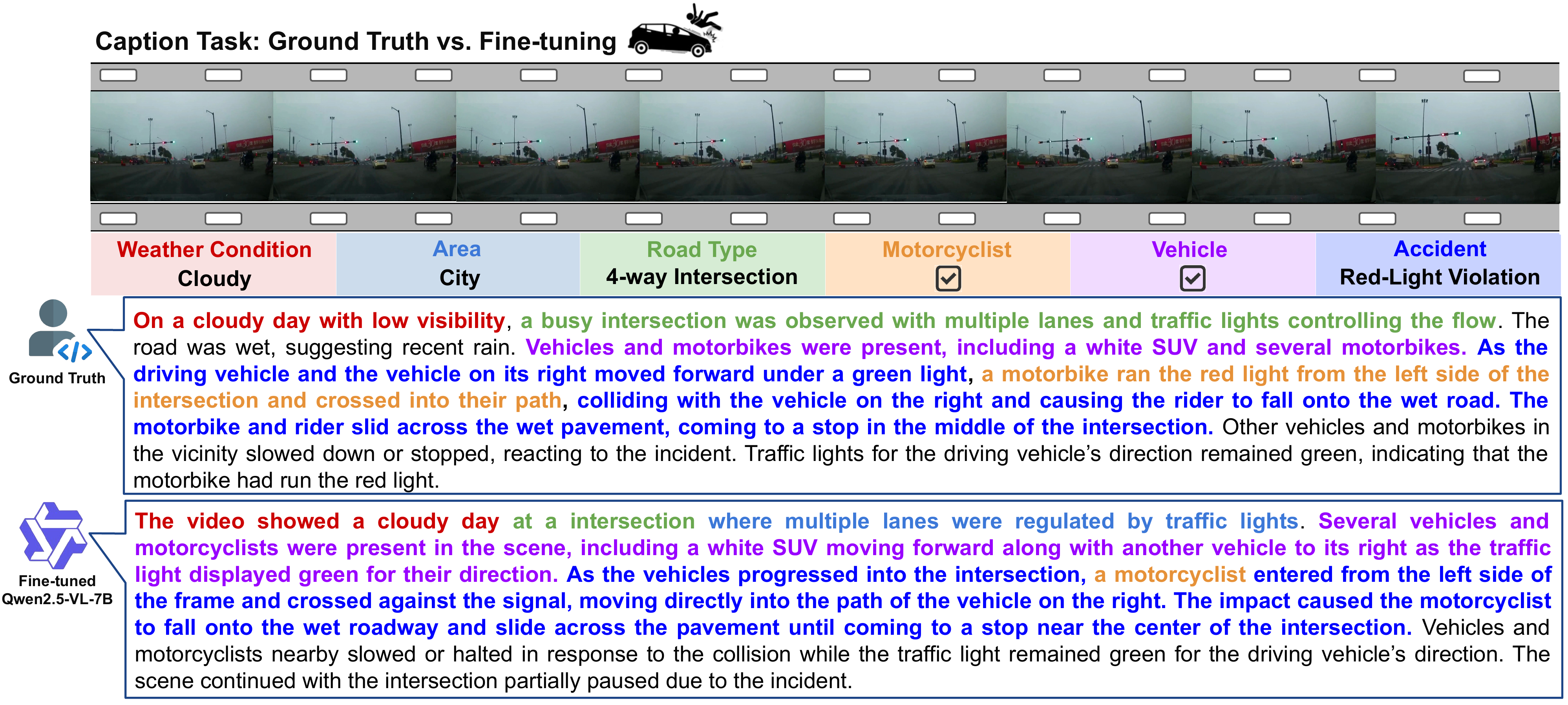}
	\end{minipage}
    \caption{Examples of accident cases across the five major categories in RoadSafe365.}
	\label{fig:vqa} 
\end{figure*}

\begin{figure*}[htb]
	\begin{minipage}[b]{1\textwidth}
		\centering		\includegraphics[width=0.98\textwidth]{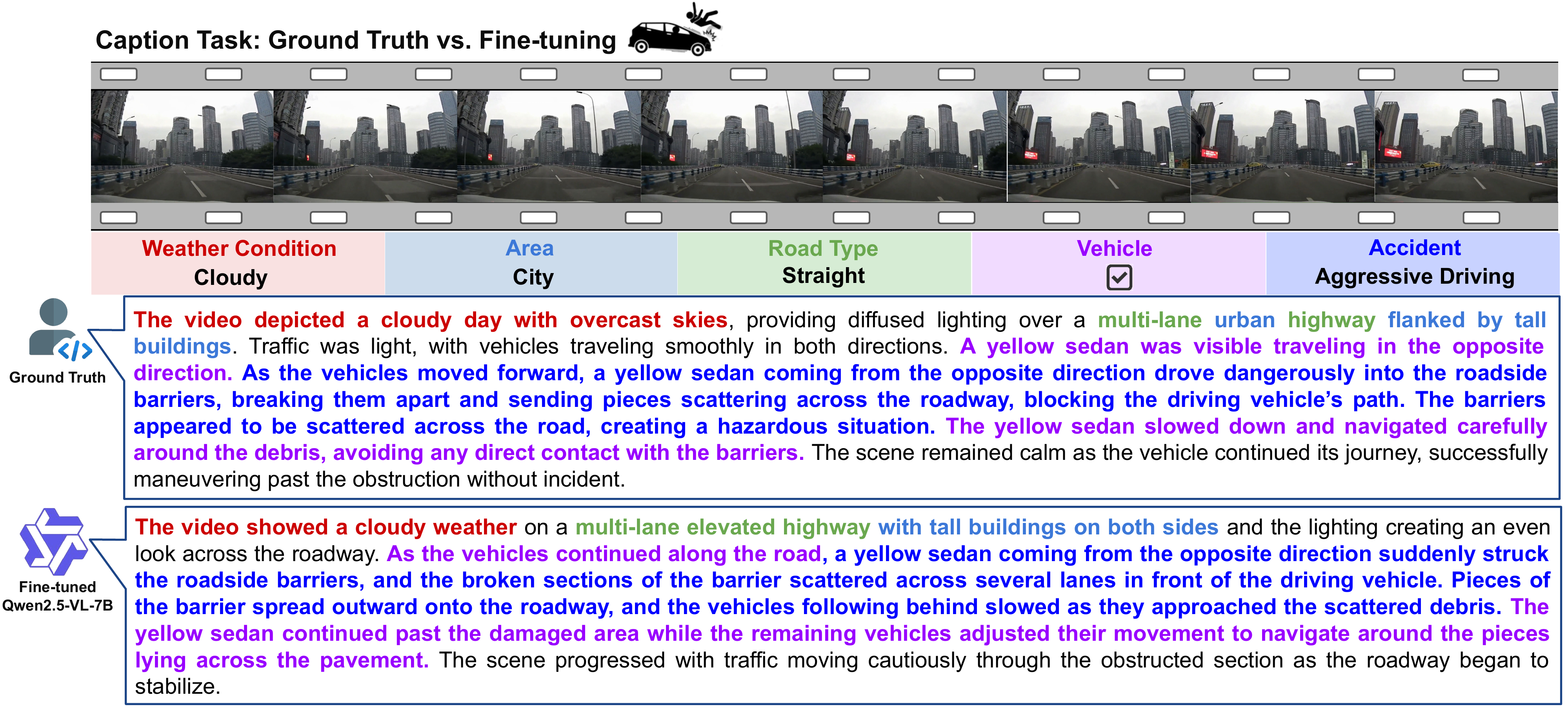}
	\end{minipage}
    \caption{Examples of accident cases across the five major categories in RoadSafe365.}
	\label{fig:vqa} 
\end{figure*}

\begin{figure*}[htb]
	\begin{minipage}[b]{1\textwidth}
		\centering		\includegraphics[width=0.98\textwidth]{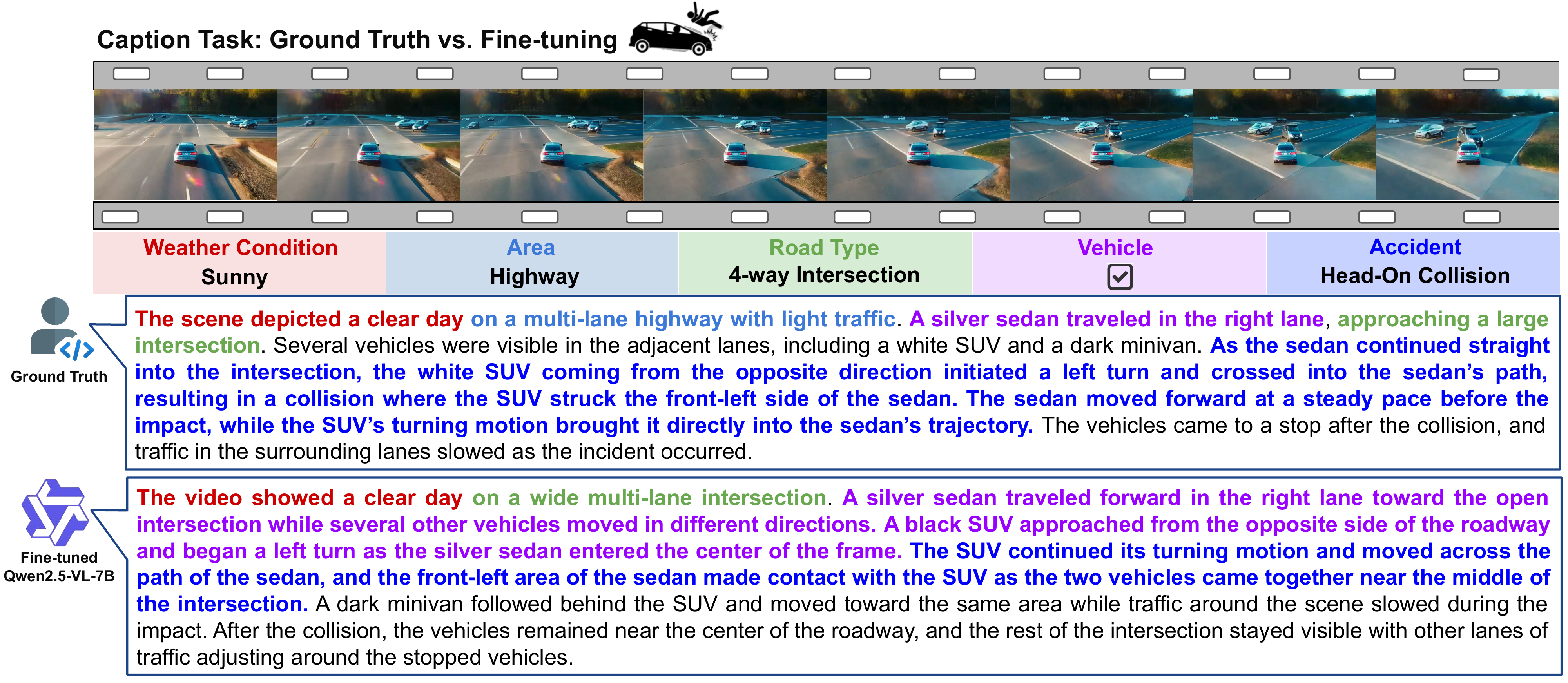}
	\end{minipage}
    \caption{Examples of synthetic accident scenes from \name-Synthetic, created from Dreamland in third-person view.}
	\label{fig:vqa} 
\end{figure*}

\begin{figure*}[htb]
	\begin{minipage}[b]{1\textwidth}
		\centering		\includegraphics[width=0.98\textwidth]{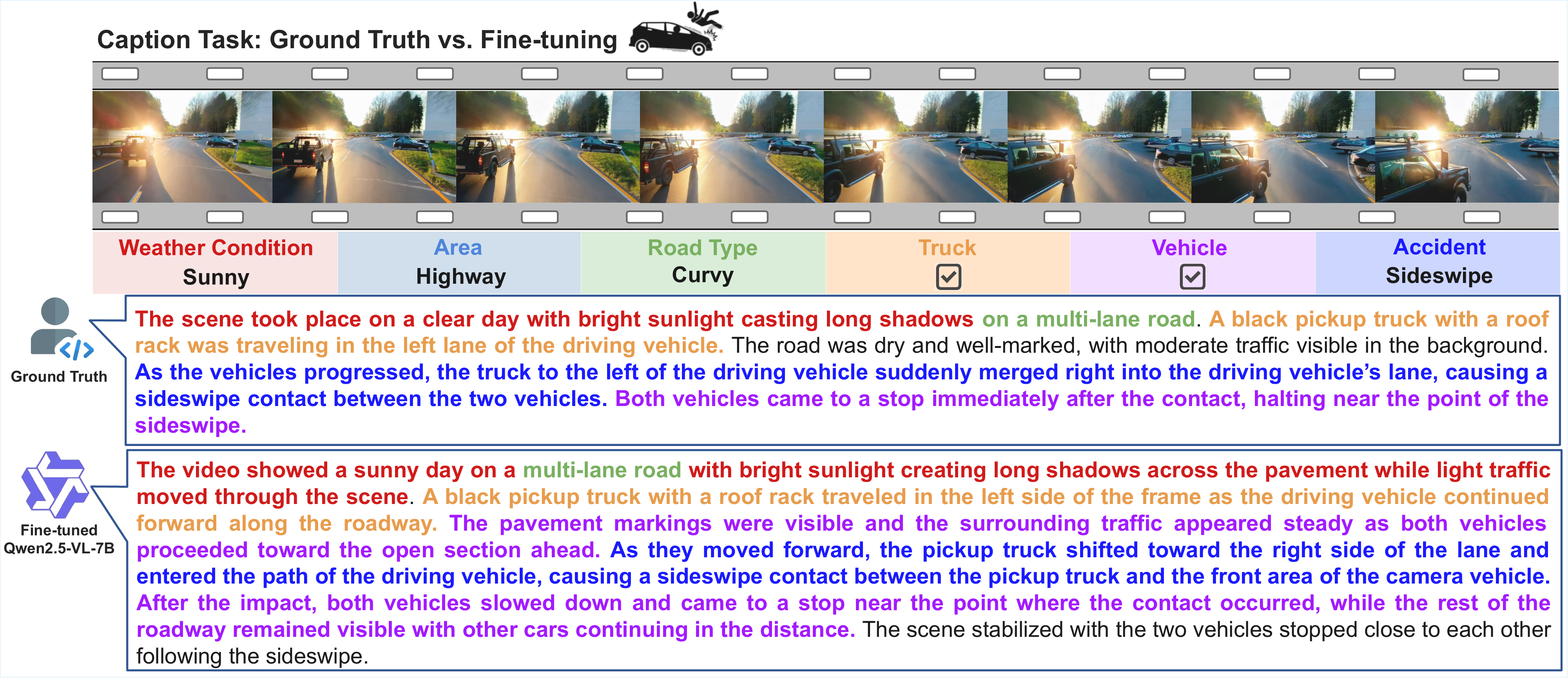}
	\end{minipage}
    \caption{Examples of synthetic accident scenes from \name-Synthetic, created from Dreamland~\cite{mo2025dreamland} in first-person view.}
	\label{fig:vqa} 
\end{figure*}



\end{document}